\documentclass[10pt,journal,compsoc]{IEEEtran}
\usepackage{times}
\usepackage{epsfig}
\usepackage{graphicx}
\usepackage{amsmath}
\usepackage{amssymb}
\usepackage{microtype}
\usepackage{changepage}
\usepackage{color}
\usepackage{url}
\usepackage{adjustbox}
\usepackage{booktabs}
\usepackage{array}
\usepackage{balance} 
\usepackage{paralist} 
\usepackage[caption=false]{subfig}
\newcolumntype{M}{>{\centering\arraybackslash}m{0.6in}}
\usepackage{multirow}
\usepackage{rotating}
\usepackage{cleveref}
\usepackage{siunitx}

\newcommand{\rulesep}{\unskip\ \vrule\ }

\def\recipe{Recipe1M}
\def\recipeplus{Recipe1M\raisebox{0.25ex}{+}}

\ifCLASSOPTIONcompsoc
  \usepackage[nocompress]{cite}
\else
  \usepackage{cite}
\fi

\usepackage{url}

\begin{document}

\title{\textcolor{black}{\recipeplus{}}: A Dataset for Learning Cross-Modal Embeddings for Cooking Recipes and Food Images}

\author{%
Javier Mar\'in$^1$*,
\and
Aritro Biswas$^1$*\thanks{*contributed equally.},
\and
Ferda Ofli$^2$, 
\and
Nicholas Hynes$^1$,
\and
Amaia Salvador$^3$,
\and
Yusuf Aytar$^1$,
\and
Ingmar Weber$^2$,
\and
Antonio Torralba$^1$
\\
$^1$Massachusetts Institute of Technology 
$^2$Qatar Computing Research Institute, HBKU \\
$^3$Universitat Polit\`ecnica de Catalunya \\
\small \{abiswas,nhynes\}@mit.edu, \{jmarin,yusuf,torralba\}@csail.mit.edu, amaia.salvador@upc.edu, \{fofli,iweber\}@hbku.edu.qa
}
\markboth{IEEE TRANSACTIONS ON PATTERN ANALYSIS AND MACHINE INTELLIGENCE, VOL. X, NO. X, MONTH YEAR}
{Mar\'in \MakeLowercase{\textit{et al.}}: Cross-Modal Embeddings for Cooking Recipes and Food Images}

\IEEEtitleabstractindextext{%
\begin{abstract}
In this paper, we introduce \textcolor{black}{\recipeplus{}}, a new large-scale, structured corpus of over one million cooking recipes and 13 million food images. As the largest publicly available collection of recipe data, \textcolor{black}{\recipeplus{}} affords the ability to train high-capacity models on aligned, multimodal data. Using these data, we train a neural network to learn a joint embedding of recipes and images that yields impressive results on an image-recipe retrieval task. Moreover, we demonstrate that regularization via the addition of a high-level classification objective both improves retrieval performance to rival that of humans and enables semantic vector arithmetic.
We postulate that these embeddings will provide a basis for further exploration of the \textcolor{black}{\recipeplus{}} dataset and food and cooking in general. Code, data and models are publicly available\footnote{\url{http://im2recipe.csail.mit.edu}}.
\end{abstract}

}

\maketitle
\IEEEdisplaynontitleabstractindextext
\IEEEpeerreviewmaketitle

\IEEEraisesectionheading{\section{Introduction}\label{sec:introduction}}

\IEEEPARstart{T}{here} are few things so fundamental to the human experience as food. Its consumption is intricately linked to our health, our feelings and our culture. Even migrants starting a new life in a foreign country often hold on to their ethnic food longer than to their native language. Vital as it is to our lives, food also offers new perspectives on topical challenges in computer vision like finding representations that are robust to occlusion and deformation (as occur during ingredient processing).

The profusion of online recipe collections with user-submitted photos presents the possibility of training machines to automatically understand food preparation by jointly analyzing ingredient lists, cooking instructions and food images. Far beyond applications solely in the realm of culinary arts, such a tool may also be applied to the plethora of food images shared on social media to achieve insight into the significance of food and its preparation on public health \cite{garimellaetal16chi} and cultural heritage \cite{mejovaetal16icwsm}. Developing a tool for automated analysis requires large and well-curated datasets.

\begin{figure}[t]
    \centering
    \centerline{
        \includegraphics[width=1\linewidth]{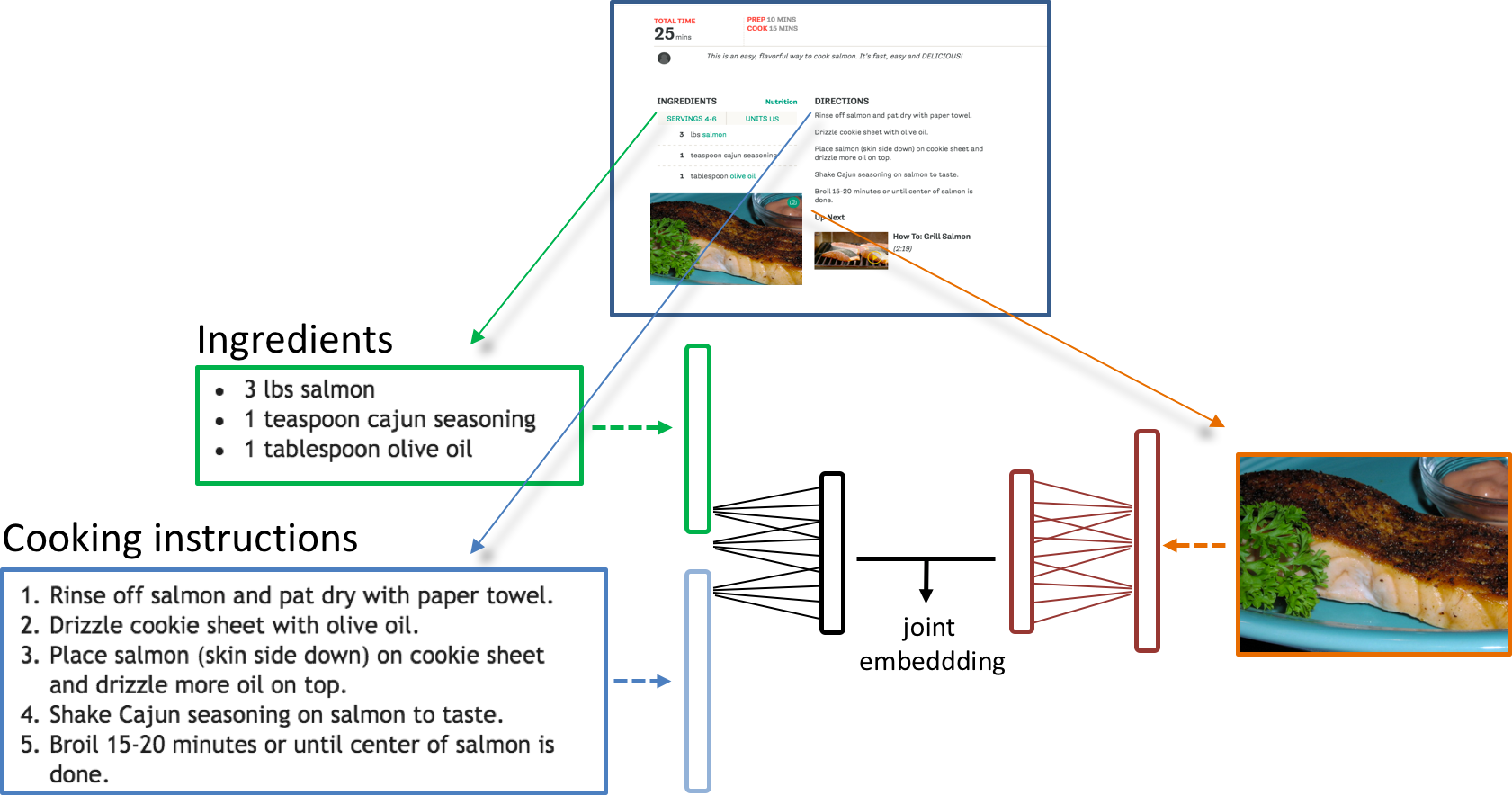}
    }
    \caption{\textbf{Learning cross-modal embeddings} from recipe-image pairs collected from online resources. These embeddings enable us to achieve in-depth understanding of food from its ingredients to its preparation.}
    \label{fig:overview}    
\end{figure}

The emergence of massive labeled datasets~\cite{russakovsky2015imagenet,zhou2014learning} and deeply-learned  representations~\cite{krizhevsky2012imagenet,simonyan2014very,he2015deep} have redefined the state-of-the-art in object recognition and scene classification. Moreover, the same techniques have enabled progress in new domains like dense labeling and image segmentation. Perhaps the introduction of a new large-scale food dataset--complete with its own intrinsic challenges--will yield a similar advancement of the field. For instance, categorizing an ingredient's state (e.g., sliced, diced, raw, baked, grilled, or boiled) provides a unique challenge in attribute recognition--one that is not well posed by existing datasets. Furthermore, the free-form nature of food suggests a departure from the concrete task of classification in favor of a more nuanced objective that integrates variation in a recipe's structure. Hence, we argue that food images must be analyzed together with accompanying recipe ingredients and instructions in order to acquire a comprehensive understanding of ``behind-the-scene'' cooking process as illustrated in Fig.~\ref{fig:overview}.

Existing work, however, has focused largely on the use of medium-scale image datasets for performing food categorization. For instance, Bossard et al.~\cite{bossard2014food} introduced the Food-101 visual classification dataset and set a baseline of 50.8\% accuracy. Even with the impetus for food image categorization, subsequent work by~\cite{liu2016deepfood},~\cite{Myers2015im2calories} and~\cite{ofli2017saki} could only improve this result to 77.4\%, 79\% and 80.9\%, respectively, which indicates that the size of the dataset may be the limiting factor. Although Myers et al.~\cite{Myers2015im2calories} built upon Food-101 to tackle the novel challenge of estimating a meal's energy content, the segmentation and depth information used in their work are not made available for further exploration.

In this work, we address data limitations by introducing the large-scale \recipeplus{} dataset which contains one million structured cooking recipes and their images. Additionally, to demonstrate its utility, we present the im2recipe retrieval task which leverages the full dataset--images and text--to solve the practical and socially relevant problem of demystifying the creation of a dish that can be seen but not necessarily described. To this end, we have developed a multimodal neural model which jointly learns to embed images and recipes in a common space which is semantically regularized by the addition of a high-level classification task. The performance of the resulting embeddings is thoroughly evaluated against baselines and humans, showing remarkable improvement over the former while faring comparably to the latter. With the release of \recipeplus{}, we hope to spur advancement on not only the im2recipe task but also heretofore unimagined objectives which require a deep understanding of the domain and its modalities.

\subsection{Related Work}
Since we presented our initial work on the topic back in 2017~\cite{ASalvador:CVPR17}, several related studies have been published and we feel obliged to provide a brief discussion about them. 

Herranz et al.~\cite{HerranzL:arXiv18}, besides providing a detailed description on recent work focusing on food applications, propose an extended multimodal framework that relies on food imagery, recipe and nutritional information, geolocation and time, restaurant menus and food styles. In another study, Min et al.~\cite{MinW:ACMMM17} present a multi-attribute theme modeling (MATM) approach that incorporates food attributes such as cuisine style, course type, flavors or ingredient types. Then, similar to our work, they train a multimodal embedding which learns a common space between the different food attributes and the corresponding food image. Most interesting applications of their model include flavor analysis, region-oriented food summary, and recipe recommendation. In order to build their model, they collect all their data from a single data source, i.e., Yummly\footnote{\url{https://www.yummly.com/}}, which is an online recipe recommendation system.

\textcolor{black}{In another interesting study, Chang et al.~\cite{ChangM:CHI18} focus on analyzing several possible preparations of a single dish, like ``chocolate chip cookie.'' The authors design an interface that allows users to explore the similarities and differences between such recipes by visualizing the structural similarity between recipes as points in a space, in which clusters are formed according to how similar recipes are. Furthermore, they examine how cooking instructions overlap between two recipes to measure recipe similarity. Our work is of a different flavor, as the features they use to measure similarity are manually picked by humans, while ours are automatically learned by a multimodal network.}

\textcolor{black}{Getting closer to the information retrieval domain, Engilberge et al.~\cite{EngilbergeM:CVPR18} examine the problem of retrieving the best matching caption for an image. In order to do so, they use neural networks to create embeddings for each caption, and retrieve the one whose embedding most closely matches the embedding of the original image. In our work, we aim to also use embeddings to retrieve the recipe matching an image, or vice versa. However, since our domain involves cooking recipes while theirs only involves captions, we account for two separate types of text -- ingredients and cooking instructions -- and combine them in a different way in our model.}

\textcolor{black}{Alternatively, Chen et al.~\cite{ChenJ:ACMMMM17} study the task of retrieving a recipe matching a corresponding food image in a slightly different way. The authors find that, although ingredient composition is important to the appearance of food, other attributes such as the manner of cutting and manner of cooking ingredients also play a role in forming the food's appearance. Given a food image, they attempt to predict ingredient, cutting and cooking attributes, and use these predictions to help retrieve the correct corresponding recipe. With our model, we attempt to retrieve the recipe directly, without first predicting attributes like ingredients, cutting and cooking attributes, separately. Furthermore, along with retrieving the recipe matching an image, our model also allow to retrieve the image matching a corresponding recipe.}

\textcolor{black}{The two most relevant studies to the current one are presented in~\cite{ChenJ:ACMMMM18} and~\cite{CarvalhoM:SIGIR18}. Different from our work, Chen et al.~\cite{ChenJ:ACMMMM18} approach the image-to-recipe retrieval problem from the perspective of attention modeling where they incorporate word-level and sentence-level attentions into their recipe representation and align them with the corresponding image representation such that both text and visual features have high similarity in a multi-dimensional space. Another difference is that they employ a rank loss instead of a pairwise similarity loss as we do. These improvements effectively lead to slight performance increases in both image-to-recipe and recipe-to-image retrieval tasks.}

\textcolor{black}{On the other hand, building upon the same network architecture as in our original work~\cite{ASalvador:CVPR17} to represent the image and text (recipe) modalities, Carvalho et al.~\cite{CarvalhoM:SIGIR18} improve our initial results further by proposing a new objective function that combines retrieval and classification tasks in a double-triplet learning scheme. This new scheme captures both instance-based (i.e., fine-grained) and semantic-based (i.e., high-level) structure simultaneously in the latent space since the semantic information is directly injected into the cross-modal metric learning problem as opposed to our use of classification task as semantic regularization. Additionally, 
they follow an adaptive training strategy to account for the vanishing gradient problem of the triplet losses 
and use the MedR score instead of the original loss in the validation phase for early stopping. We also find that using the MedR score as the performance measure in the validation phase is more stable. However, our work is orthogonal to both of these studies, i.e., 
their performances can be further improved with the use of our expanded dataset and the quality of their embeddings can be further explored with various arithmetics presented in this submission.}


The rest of the paper is organized as follows. In Section~\ref{sec:dataset}, we introduce our large-scale, multimodal cooking recipe dataset and provide details about its collection process. We describe our recipe and image representations in Section~\ref{sec:embeddings} and present our neural joint embedding model in Section~\ref{sec:joint_embedding}. Then, in Section~\ref{sec:semantic}, we discuss our semantic regularization approach to enhance our joint embedding model. In Section~\ref{sec:experiments}, we present results from our various experiments and conclude the paper in Section~\ref{sec:conclusion}.

\section{Dataset}
\label{sec:dataset}


Due to their complexity, textually and visually, (e.g., ingredient-based variants of the same dish, different presentations, or multiple ways of cooking a recipe), understanding food recipes demands a large, general collection of recipe data. Hence, it should not be surprising that the lack of a larger body of work on the topic could be the result of missing such a collection. To our knowledge, practically all the datasets publicly available in the research field either contain only categorized images \cite{Myers2015im2calories,bossard2014food,kawanoyanai15mta,msg} or simply recipe text \cite{kusmierczyketal16hypertext}. Only recently have a few datasets been released that include both recipes and images. For instance, Wang et al.~\cite{wangetal15icme} released a multimodal food dataset which has 101k images divided equally among 101 food categories; the recipes for each are however raw HTML. In a later work, Chen and Ngo~\cite{VireoFood172} presented a dataset containing 110,241 images annotated with 353 ingredient labels and 65,284 recipes, each with a brief introduction, ingredient list, and preparation instructions. Of note is that the dataset only contains recipes for Chinese cuisine. 

Although the aforementioned datasets constitute a large step towards learning richer recipe representations, they are still limited in either generality or size. As the ability to learn effective representations is largely a function of the quantity (especially when learning features using deep architectures) and quality of the available data, we create and release publicly a new, large-scale corpus of structured recipe data that includes over 1M recipes and 13M images. In comparison to the current largest datasets in this domain, the \recipeplus{} includes twice as many recipes as \cite{kusmierczyketal16hypertext} and 130 times as many images as \cite{VireoFood172}.

We created the \recipeplus{} dataset in two phases. In the first phase, we collected a large dataset of cooking recipes paired with food images, all scraped from a number of popular cooking websites, which resulted in more than 1M cooking recipes and 800K food images (i.e., \recipe{}~\cite{ASalvador:CVPR17}). Then, in the second phase, we augmented each recipe in this initial collection with food images downloaded from the Web using a popular image search engine, which amounted to over 13M food images after cleaning and removing exact-and-near duplicates. In the following subsections, we elaborate further on these data collection phases, outline how the dataset is organized, and provide analysis of its contents.

\subsection{Data Collection from Recipe Websites}
\label{ssec:dataset_collection}

The recipes were scraped from over two dozen popular cooking websites and processed through a pipeline that extracted relevant text from the raw HTML, downloaded linked images, and assembled the data into a compact JSON schema in which each datum was uniquely identified.
As part of the extraction process, excessive whitespace, HTML entities, and non-ASCII characters were removed from the recipe text.
Finally, after removing duplicates and near-matches (constituting roughly 2\% of the original data), the retained dataset contained over 1M cooking recipes and 800K food images (i.e., \recipe{}~\cite{ASalvador:CVPR17}). Although the resulting dataset is already larger than any other dataset in this particular domain (i.e., includes twice as many recipes as \cite{kusmierczyketal16hypertext} and eight times as many images as \cite{VireoFood172}), the total number of images is not yet at the same scale as the largest publicly available datasets such as ImageNet~\cite{russakovsky2015imagenet} and Places~\cite{Zhou2018places_pami}, which contain tens of millions of images, in the computer vision community. Therefore, in the next phase, we aimed to extend the initial collection of images by querying for food images through an image search engine.

\subsection{Data Extension using Image Search Engine}
\label{ssec:dataset_extension}

\begin{figure}[!t]
    \centering
    \centerline{
        \includegraphics[width=1\linewidth]{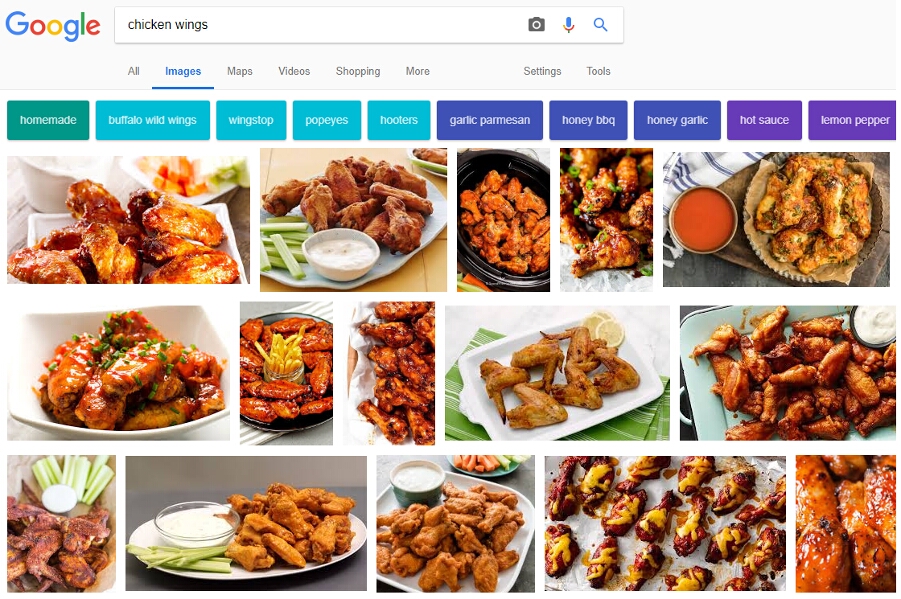}
    }
    \caption{\textbf{Google image search results.} The query used is \textit{chicken wings}.}
    \label{fig:chickenwings}
        \vspace{-0.5em}
\end{figure}

Thanks to the latest technological infrastructure advances, half the population of the entire world have become Internet users\footnote{\url{https://www.internetworldstats.com/stats.htm}}. Online services ranging from social networks to simple websites have grown into data containers where users share images, videos, or documents. Companies like Google, Yahoo, and Microsoft, among others, offer public search engines that go through the entire Internet looking for websites, videos, images and any other type of content that matches a text query (some of them also support image queries). Looking at the search results for a given recipe title (e.g., ``chicken wings'') in Fig.~\ref{fig:chickenwings}, one can say that the retrieved images are generally of very good quality. We also observed during the first phase of data collection from recipe websites that users were often using images from other recipes of the same dish (sometimes with slight differences) to visually describe theirs. Motivated by these insights, we downloaded a large amount of images using as queries the recipe titles collected from the recipe websites in the first phase.

\smallskip\noindent{\bf Data Download.} We targeted collecting 50M images, i.e., 50 images per recipe in the initial collection. In order to amass such a quantity of images, we chose the Google search engine. As mentioned before, we used the title of each recipe as a query. Out of the Google search results, we selected the top 50 retrieved images and stored locally their image URLs.
For this task, we used publicly available Python libraries on ten servers in parallel for several days. Then, to download images simultaneously, we made use of Aria2\footnote{\url{https://aria2.github.io/}}, a publicly available download utility. In the end, we managed to download over 47M images as some of the image URLs either were corrupted or did not exist any more.

\begin{table}[!t]
    \centering
	\caption{\textbf{Dataset sizes.} Number of recipes and images in training, validation and test sets of each dataset.}    
		\begin{tabular}{lcccc}
        	\toprule
        	& & \recipe{} & intersection & \recipeplus{}\\
        	\midrule
			Partition & \# Recipes & \# Images & \# Images & \# Images\\
			\midrule
			Training & 720,639 & 619,508 & 493,339 & 9,727,961 \\
			Validation & 155,036 & 133,860 & 107,708 & 1,918,890 \\
			Test  & 154,045 & 134,338 & 115,373 & 2,088,828\\
            \midrule
            Total & 1,029,720 & 887,706 & 716,480 & 13,735,679\\
            \bottomrule
		\end{tabular}
\label{tvtdatastats}
\end{table}

\smallskip\noindent{\bf Data Consolidation.} One of the first tasks, besides removing corrupted or wrong format images, was eliminating the duplicate images. For this task, we simply used a pre-trained ResNet18~\cite{he2015deep} as a feature extractor (by removing its last layer for classification) and computed pairwise euclidean distances between the collected images. During this cleanse process, we combined the initial set of images collected from recipe websites and the new ones collected via Google image search. After this first stage, we removed over 32M duplicate images (those with an euclidean distance of 0). We only kept one representative for each duplicate cluster.
Later, we visually inspected the remaining images and realized that a significant amount of them were still either duplicates or near-duplicates. The main reason we could not detect some of these duplicates in the first stage was due to compression or rescaling operations applied to the images, which cause slight modifications to their feature representation. By using distances between them, and removing those that were close enough, we managed to eliminate these duplicates. Near-duplicates, instead, were due to distortions (i.e., aspect-ratio changes), crops, added text into the image, and other alterations. To remove near-duplicates, after trying different strategies, we chose a harsh distance threshold between images, which meant we had to eliminate a certain amount of good examples, as well. This strategy was used between different partitions (i.e., training, test and validation). That is, we allowed near-duplicates within a partition to a certain extent (using a relaxed threshold). Additionally, we ran a face detector over the images and removed those that had a face with high confidence. Thanks to computing distances, we also found non-recipe images such as images with nutritional facts. Images containing only text were close to each other within the feature space. In order to compute the distances between images, we used C\raisebox{.25\height}{\scalebox{0.8}{++}} over Python for efficiency purposes.

\textcolor{black}{Regarding the recipes sharing the same title, we uniformly distributed the queried images for a particular non-unique title among the recipes sharing it. This helped us to avoid having different recipes with the exact same food images. In the last two paragraphs of Section 2.5, we describe an experiment performed by humans that supports the validity of spreading them uniformly.}

\textcolor{black}{In order to re-balance the dataset in terms of partitions, we slightly modified the images belonging to each partition. For a fair comparison between the \recipe{} and \recipeplus{} in our experiments, we created an \textit{intersection} version of the initial dataset, which simply contains the images that were common between both of them. One would expect \recipe{} images to be a subset of \recipeplus{} images, but due to the re-balance and the cleanse of near-duplicates, which were not done in the original \recipe{} dataset, this was no longer true. Table~\ref{tvtdatastats} shows the small differences in numbers.}

\subsection{Nutritional Information}
\label{ssec:dataset_nutritional}

The ingredient lists in the recipes scraped from the recipe websites include the ingredient, quantity and unit information altogether in a single sentence in several cases. In order to simplify the task of automatically computing the nutritional information of a recipe, we decided to encapsulate these three different fields, i.e., (i) the ingredient, (ii) the units, and (iii) the quantity, separately in the dataset structure. After identifying different type of sentences that followed the `quantity-unit-ingredient' sequence pattern in the recipe ingredient lists, we used a natural language processing toolkit\footnote{\url{http://www.nltk.org/}} to tag every single word within each of these sentences (e.g., [(`2', `CD'), (`cups', `NNS'), (`of', `IN'), (`milk', `NN')]). Every ingredient in the dataset that followed the sentence structure (e.g., `4 teaspoons of honey') of one of those we identified, was selected for further processing. We then went through the unit candidates of these sentences and chose only the measurable ones (some non-measurable units are for instance, a \textit{bunch}, a \textit{slice} or a \textit{loaf}). Table~\ref{table:units} shows the 20 different units we found. 103,152 unique recipes had measurable units and numerical quantities defined for all their ingredients. Regarding numerical quantities, these recipes contained 1,002 different ones.


\begin{table}[!t]
  \centering
    \caption{\textbf{\recipeplus{} units.} The 20 measurable units isolated in the dataset.}
    \begin{tabular}{cc}
         \toprule
         units & \\
         \midrule
 
		& \parbox[t]{5cm}{bushel, cup, dash, drop, fl. oz, g, gallon, glass, kg, liter, ml, ounce, pinch, pint, pound, quart, scoop, shot, tablespoon, teaspoon} \\
      \bottomrule
    \end{tabular}
  \label{table:units}

\end{table}

\begin{figure}[!t]
    \centering
    \centerline{
        \includegraphics[width=1\linewidth]{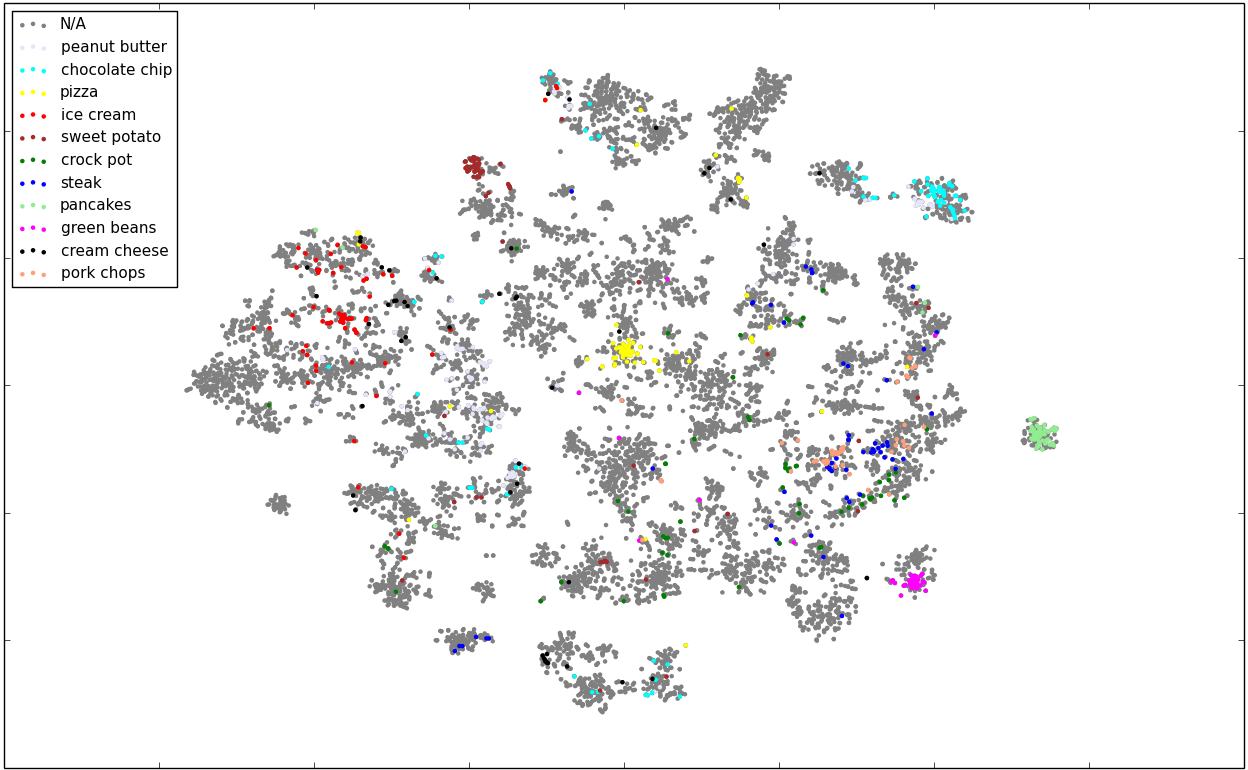}
    }
    \caption{\textbf{Embedding visualization using t-SNE}. Legend depicts the recipes that belong to the top 12 semantic categories used in our semantic regularization (see Section~\ref{sec:semantic} for more details).}
    \label{fig:t-sne}
        \vspace{-0.5em}
\end{figure}

\begin{figure}[!t]
    \centering
    \centerline{
        \includegraphics[width=1\linewidth]{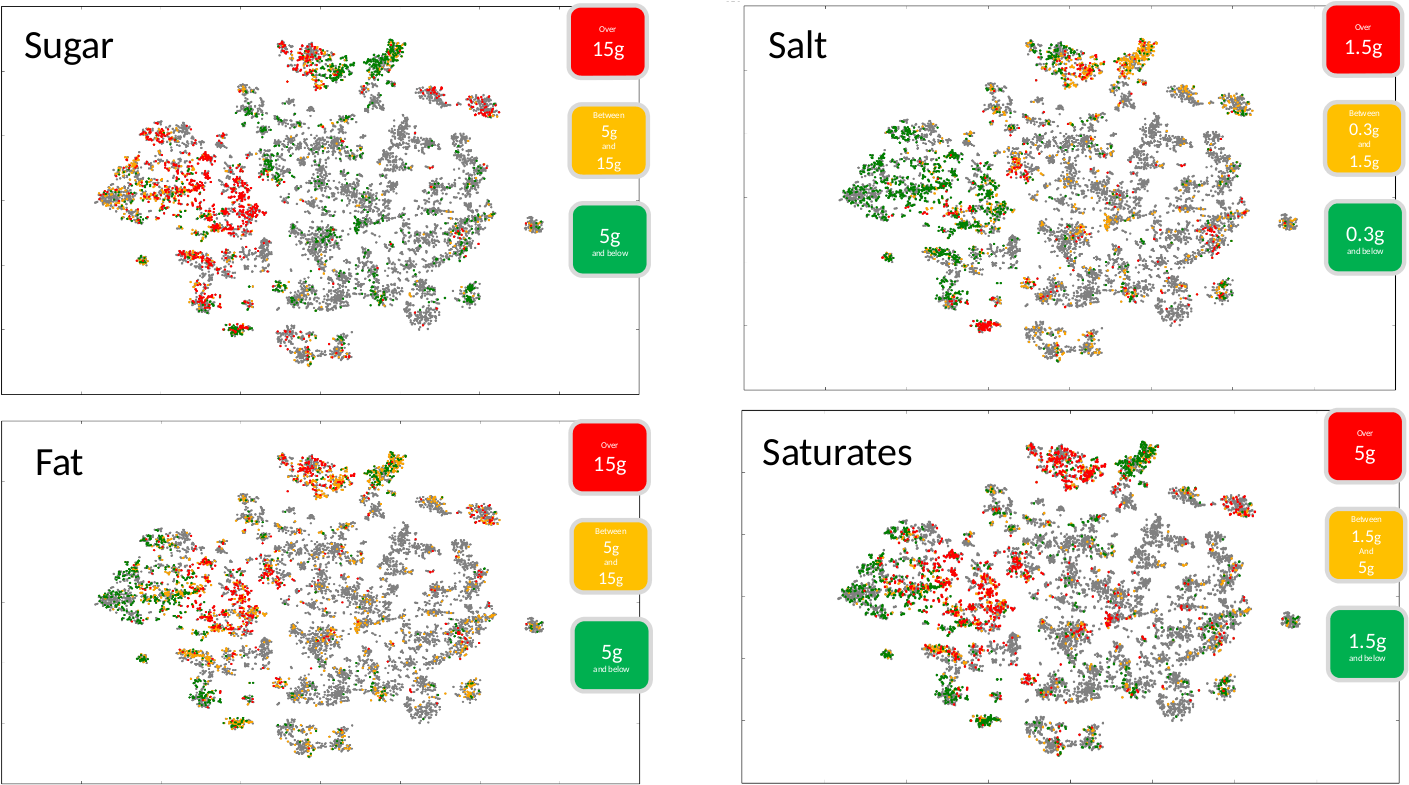}
    }
    \caption{\textbf{Healthiness within the embedding.} Recipe health is represented within the embedding visualization in terms of sugar, salt, saturates, and fat. We follow FSA traffic light system to determine how healthy a recipe is.}
    \label{fig:health}
        \vspace{-0.5em}
\end{figure}

\textcolor{black}{Once we finished the previous stage, we matched thousands of ingredient names with a publicly available nutrient database~\cite{usda:2015} assembled by the United States Department of Agriculture (USDA). This database provides the nutritional content of over 8,000 generic and proprietary-branded foods.} In order to facilitate the matching process, we first reduced the ingredient list to contain only the first word within the sentence (after removing quantities and units), obtaining a total of 6,856 unique words. Then, for each unique ingredient we picked, when available, the second word of the sentence. Due to multiple different sentences having the same first word, we did only take one example out of the possible ones. We went through each single bigram and only selected those that were food ingredients, e.g., \textit{apple juice} or \textit{cayenne pepper}. \textcolor{black}{If the second word was nonexistent, e.g., \textit{$1/2$ spoon of sugar}, or was not part of a standard ingredient name, e.g., \textit{$1$ cup of water at \SI{40}{\celsius}}, we only selected the first word, i.e., \textit{sugar} and \textit{water}, respectively.} We created a corpus of 2,057 unique different ingredients with their singular and plural versions, and, in some cases, synonyms or translations, e.g., \textit{cassava} can be also called \textit{yuca}, \textit{manioc} or \textit{mandioca}. We found ingredient names from different nationalities and cultures, such as Spanish, Turkish, German, French, Polish, American, Mexican, Jewish, Indian, Arab, Chinese or Japanese among others. Using the ingredient corpus we assigned to each ingredient sentence the closest ingredient name by simply verifying that all the words describing the ingredient name were within the original ingredient sentence. We found 68,450 recipes with all their ingredients within the corpus. The matching between the USDA database and the new assigned ingredient names, similarly as before, was done by confirming that all the words describing the ingredient name were within one of the USDA database food instances. We inspected the matching results to assure the correctness. In the end, we obtained 50,637 recipes with nutritional information (mapping example: \textit{American cheese} $\Rightarrow$ \textit{cheese, pasteurized process, American, without added vitamin d}). In Fig.~\ref{fig:t-sne}, we can see a 2D visualization of the embeddings of these recipes that also include images, using t-SNE \cite{tsne}. Recipes are shown in different colors based on their semantic category (see Section~\ref{sec:semantic}). In Fig.~\ref{fig:health}, we can see the same embedding but this time showing the same recipes on different colors depending on how healthy they are in terms of sugar, fat, saturates, and salt. We used the traffic lights\footnote{https://www.resourcesorg.co.uk/assets/pdfs/foodtrafficlight1107.pdf} definition established by the Food Standards Agency (FSA).

\begin{figure*}[!t]
  \centering
  \includegraphics[width=0.26\textwidth]{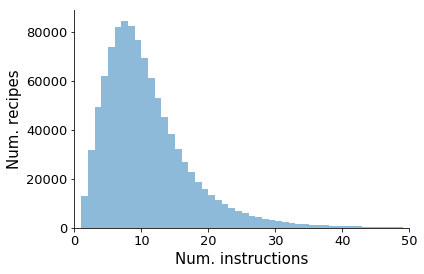}
  \includegraphics[width=0.26\textwidth]{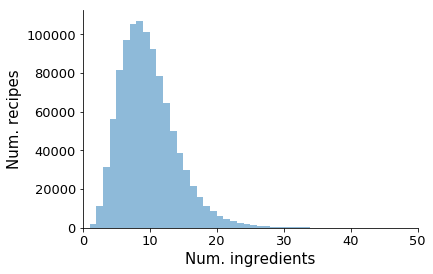}
  \includegraphics[width=0.26\textwidth]{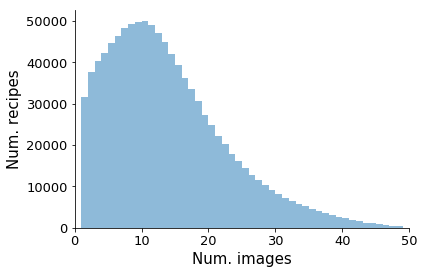}
  \raisebox{0.8em}{\includegraphics[width=0.20\textwidth]{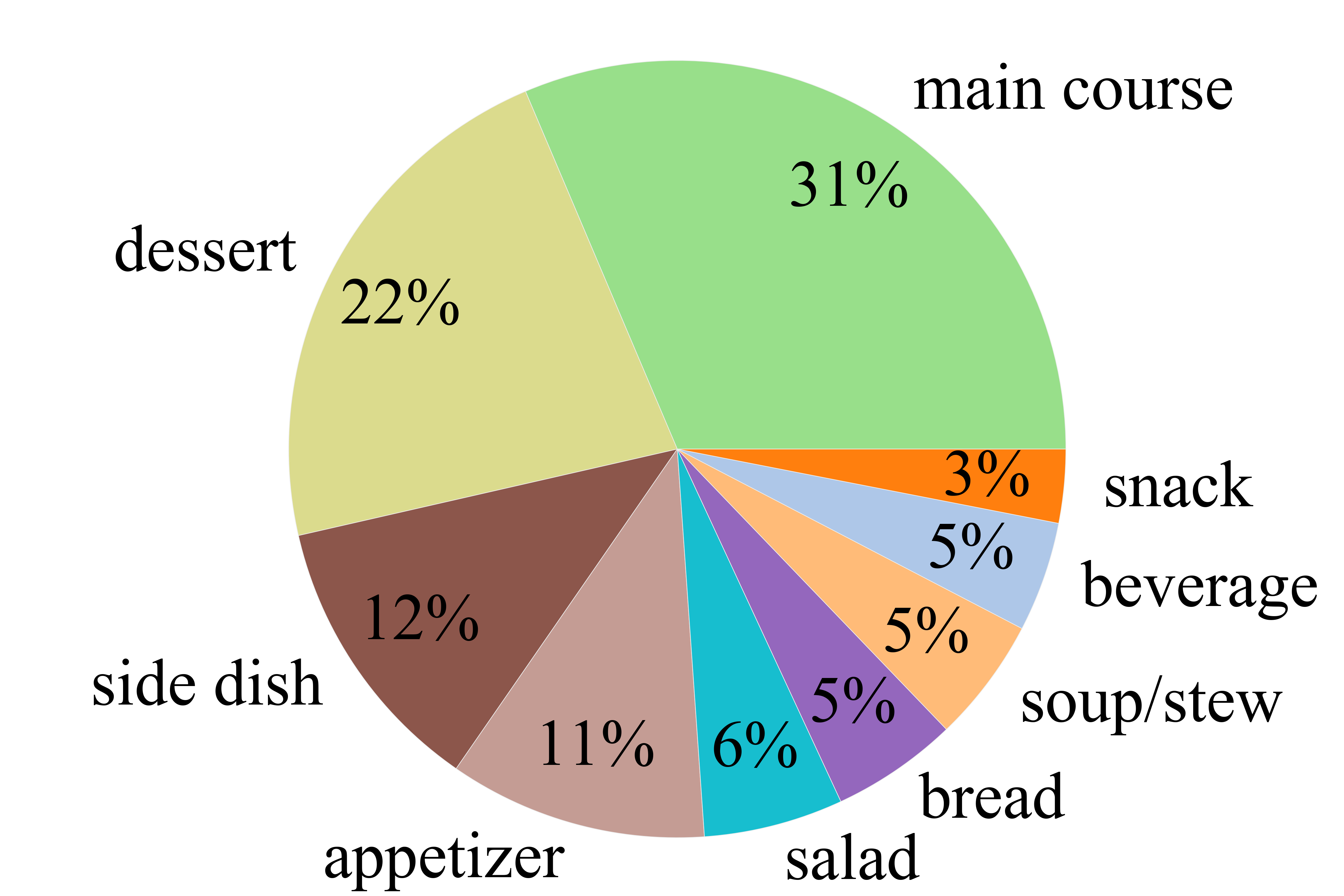}}
  \caption{\textbf{Dataset statistics.} Prevalence of course categories and number of instructions, ingredients and images per recipe in \recipeplus{}.}
  \label{fig:dsstats}
\end{figure*}

\subsection{Data Structure}
\label{ssec:dataset_structure}

The contents of the \recipe{} dataset can logically be grouped into two layers. The first layer (i.e., Layer 1) contains basic information including a title, a list of ingredients, and a sequence of instructions for preparing a dish; all of these data are provided as free text. Additional fields such as unit and quantity are also available in this layer. In cases where we were unable to extract unit and quantity from the ingredient description, these two fields were simply left empty for the corresponding ingredient. Nutritional information (i.e., total energy, protein, sugar, fat, saturates, and salt content)  is only added for those recipes that contained both units and quantities as described in Section~\ref{ssec:dataset_nutritional}. FSA traffic lights are also available for such recipes. The second layer (i.e., Layer 2) builds upon the first layer and includes all images with which the recipe is associated--these images are provided as RGB in JPEG format. Additionally, a subset of recipes are annotated with course labels (e.g., appetizer, side dish, or dessert), the prevalence of which are summarized in Fig.~\ref{fig:dsstats}. For \recipeplus{}, we provide same Layer 1 as described above with different partition assignments and Layer 2 including the 13M images.


\subsection{Analysis}
\label{ssec:analysis}

\recipe{} (hence \recipeplus{}) includes approximately 0.4\% duplicate recipes and, excluding those duplicate recipes, 20\% of recipes have non-unique titles but symmetrically differ by a median of 16 ingredients. 0.2\% of recipes share the same ingredients but are relatively simple (e.g., spaghetti, or granola), having a median of six ingredients. Approximately half of the recipes did not have any images in the initial data collection from recipe websites. However, after the data extension phase, only around 2\% of the recipes are left without any associated images. Regarding the experiments, we carefully removed any exact duplicates or recipes sharing the same image in order to avoid overlapping between training and test sets. As detailed earlier in Table~\ref{tvtdatastats}, around 70\% of the data is labeled as training, and the remainder is split equally between the validation and test sets. During the dataset extension, as we mentioned earlier, we also created an \textit{intersection} dataset in order to have a fair comparison of the experimental results on both the initial and the extended versions of the dataset.

According to Fig.~\ref{fig:dsstats}, the average recipe in the dataset consists of nine ingredients which are transformed over the course of ten instructions. One can also observe that the distributions of data are heavy tailed. For instance, 
of the 16k ingredients identified as unique (in terms of phrasing), 
only 4,000 account for 95\% of occurrences. At the low end of instruction count--particularly those with one step--one will find the dreaded \textit{Combine all ingredients}. 
At the other end are lengthy recipes and ingredient lists associated with recipes that include sub-recipes.

A similar issue of outliers exists also for images: as several of the included recipe collections curate user-submitted images, popular recipes like chocolate chip cookies have orders of magnitude more images than the average.
Notably, 
the number of unique recipes that came with associated food images in the initial data collection phase was 333K, whilst after the data extension phase, this number reached to more than 1M recipes.
On average, the \recipeplus{} dataset contains 13 images per recipe whereas \recipe{} has less than one image per recipe, 0.86 to be exact. Fig.~\ref{fig:dsstats} also depicts the images vs recipes histogram for \recipeplus{}, where over half million recipes contain more than 12 images each.

\textcolor{black}{To evaluate further the quality of match between the queried images and the recipes, we performed an experiment on Amazon Mechanical Turk (AMT) platform\footnote{\url{http://mturk.com}}. We randomly picked 3,455 recipes, containing at most ten ingredients and ten instructions, from the pool of recipes with non-unique titles. Then, for each one of these recipes, we showed AMT workers a pair of images and asked them to choose which image, A or B, was the best match for the corresponding recipe. The workers also had the options of selecting `both images' or `none of them'. Image A and image B were randomly chosen; one from the original recipe (i.e., Recipe1M) images and the other one from the queried images collected during the dataset expansion for the corresponding recipe title. We also changed the order of image A and image B randomly. We explicitly asked the workers to check all the ingredients and instructions. Only master workers were selected for this experiment.} 

\textcolor{black}{Out of 3,455 recipes, the workers chose 971 times the original recipe image (28.1\%); 821 times the queried one (23.8\%); 1581 times both of them (45.8\%); and 82 times none of them (2.4\%). Given the difference between the original recipe image vs.\ the queried image is less than 5\%, these results show that the extended dataset is not much noisier than the original \recipe{}.}


\section{Learning Embeddings}
\label{sec:embeddings}

In this section, we describe our neural joint embedding model. Here, we utilize the paired (recipe and image) data in order to learn a common embedding space as illustrated in Fig.~\ref{fig:overview}. Next, we discuss recipe and image representations, and then, we describe our neural joint embedding model that builds upon recipe and image representations.

\subsection{Representation of Recipes}
\label{ssec:embeddings_recipe}

There are two major components of a recipe: its ingredients and cooking instructions. We develop a suitable representation for each of these components.

\smallskip\noindent{\bf Ingredients.} Each recipe contains a set of ingredient text as shown in Fig.~\ref{fig:overview}. For each ingredient we learn an ingredient level word2vec~\cite{word2vec} representation. In order to do so, the actual ingredient names are extracted from each ingredient text. For instance in ``2 tbsp of olive oil" the \emph{olive\_oil} is extracted as the ingredient name and treated as a single word for word2vec computation.
The initial ingredient name extraction task is solved by a bi-directional LSTM that performs logistic regression on each word in the ingredient text. Training is performed on a subset of our training set for which we have the annotation for actual ingredient names. Ingredient name extraction module works with $99.5\%$ accuracy tested on a held-out set.

\begin{figure*}[!t]
    \centering
    \centerline{
        \includegraphics[width=1\linewidth]{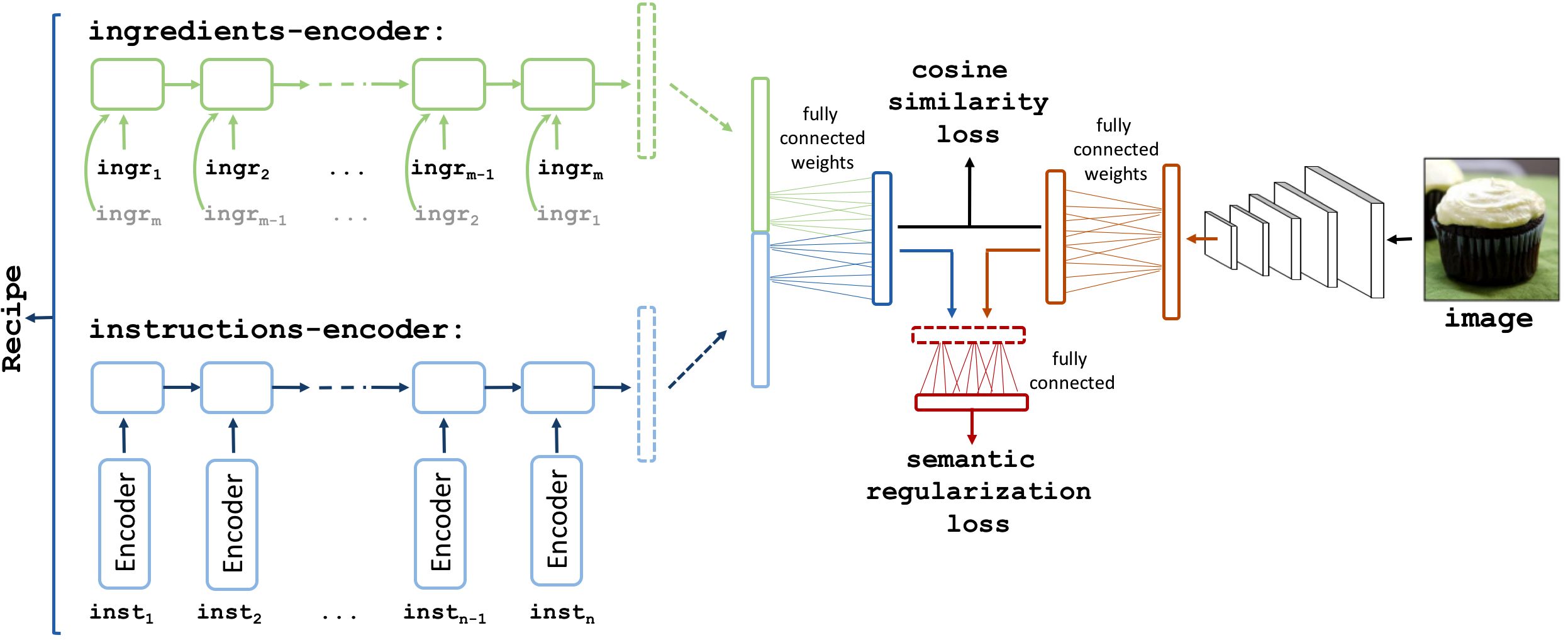}
    }
    \caption{\textbf{Joint neural embedding model with semantic regularization.} Our model learns a joint embedding space for food images and cooking recipes.}
    \label{fig:joint-embedding}
\end{figure*}

\smallskip\noindent{\bf Cooking Instructions.} Each recipe also has a list of cooking instructions. 
As the instructions are quite lengthy (averaging ${\sim}$208 words) a single LSTM is not well suited to their representation as gradients are diminished over the many time steps. Instead, we propose a two-stage LSTM model which is designed to encode a sequence of sequences. First, each instruction/sentence is represented as a \textit{skip-instructions} vector, and then, an LSTM is trained over the sequence of these vectors to obtain the representation of all instructions.
The resulting fixed-length representation is fed into to our joint embedding model (see instructions-encoder in Fig.~\ref{fig:joint-embedding}).

\begin{figure*}[ht!]
    \centering
    \centerline{
        \includegraphics[width=1\linewidth]{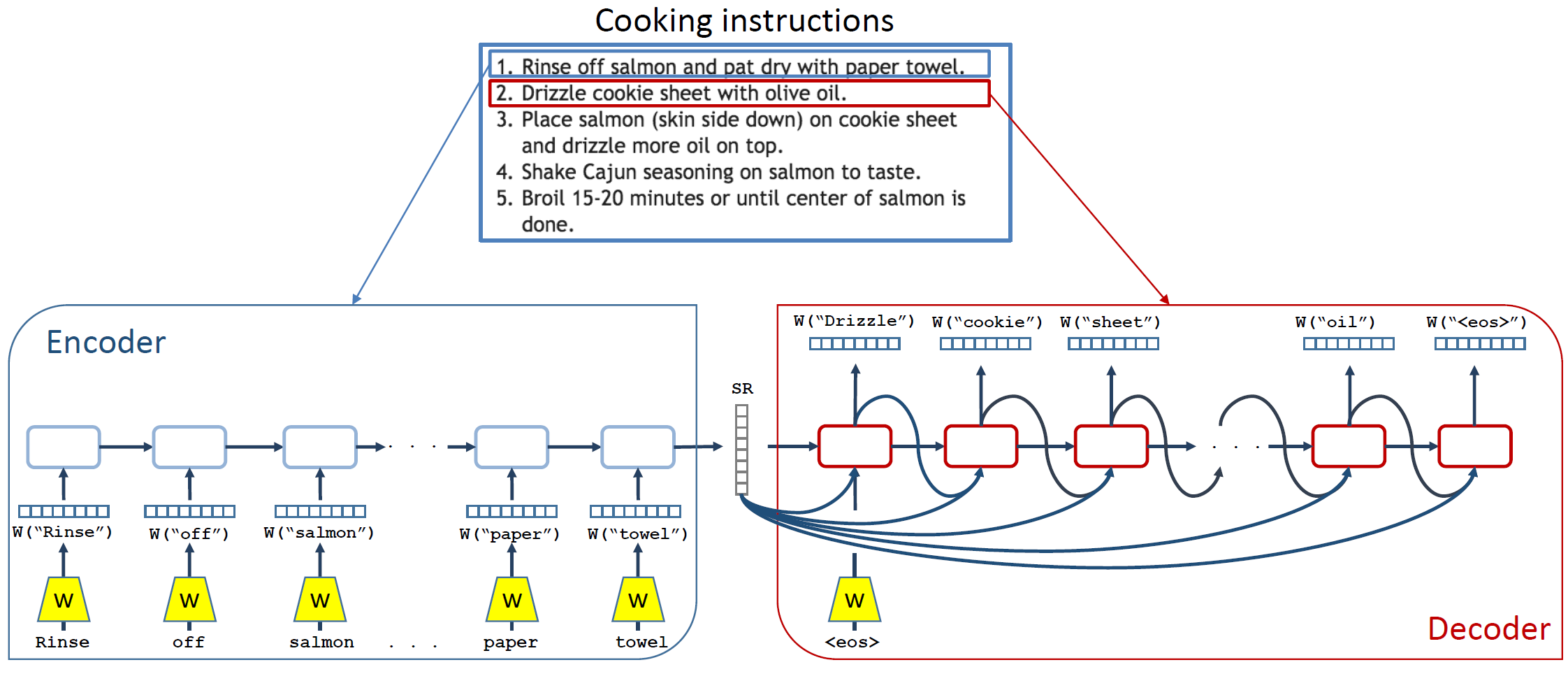}
    }
    \caption{\textbf{Skip-instructions model.} During training the encoder learns to predict the next instruction.}
    \label{fig:skip-instructions}
        \vspace{-0.5em}
\end{figure*}

\smallskip\noindent{\bf Skip-instructions.} Our cooking instruction representation, referred to as \emph{skip-instructions}, is the product of a sequence-to-sequence model~\cite{seq2seq}. Specifically, we build upon the technique of skip-thoughts~\cite{skipthoughts} which encodes a sentence and uses that encoding as context when decoding/predicting the previous and next sentences (see Fig.~\ref{fig:skip-instructions}).
Our modifications to this method include adding start- and end-of-recipe ``instructions'' and using an LSTM instead of a GRU.
In either case, the representation of a single instruction is the final output of the encoder. 
As before, this is used as the instructions input to our embedding model.

\subsection{Representation of Food Images}
\label{ssec:embeddings_image}

For the image representation we adopt two major state-of-the-art deep convolutional networks, namely VGG-16~\cite{simonyan2014very} and Resnet-50~\cite{he2015deep} models. In particular, the deep residual networks have a proven record of success on a variety of benchmarks~\cite{he2015deep}. Although \cite{simonyan2014very} suggests training very deep networks with small convolutional filters, deep residual networks take it to another level using ubiquitous identity mappings that enable training of much deeper architectures (e.g.,\ with 50, 101, or 152 layers) with better performance. We incorporate these models by removing the last softmax classification layer and connecting the rest to our joint embedding model as shown in the right side of Fig.~\ref{fig:joint-embedding}.

\section{Joint Neural Embedding}
\label{sec:joint_embedding}

Building upon the previously described recipe and image representations, we now introduce our joint embedding method. The recipe model, displayed in Fig.~\ref{fig:joint-embedding}, includes two encoders: one for ingredients and one for instructions, the combination of which are designed to learn a recipe level representation.
The ingredients encoder combines the sequence of ingredient word vectors. Since the ingredient list is an unordered set, we choose to utilize a bidirectional LSTM model, which considers both forward and backward orderings. The instructions encoder is implemented as a forward LSTM model over skip-instructions vectors. The outputs of both encoders are concatenated and embedded into a recipe-image joint space. The image representation is simply projected into this space through a linear transformation.
The goal is to learn transformations to make the embeddings for a given recipe-image pair ``close.''

Formally, assume that we are given a set of the recipe-image pairs, $(r_k, v_k)$ in which $r_k$ is the $k^{th}$ recipe and $v_k$ is the associated image. Further, let $r_k = \left(\{s_k^t\}_{t=1}^{n_k}, \{g_k^t\}_{t=1}^{m_k}\right)$, where $\{s_k^t\}_{t=1}^{n_k}$ is the sequence of $n_k$ cooking instructions, $\{g_k^t\}_{t=1}^{m_k}$ is the sequence of $m_k$ ingredient tokens.
The objective is to maximize the cosine similarity between positive recipe-image pairs, and minimize it between all non-matching recipe-image pairs, up to a specified margin. 

The ingredients encoder is implemented using a bi-directional LSTM: at each time step it takes two ingredient-word2vec representations of $g_k^t$ and $g_k^{m_k-t+1}$, and eventually, it produces the fixed-length representation $h_k^g$ for ingredients. The instructions encoder is implemented through a regular LSTM. At each time step it receives an instruction representation from the skip-instructions encoder, and finally it produces the fixed-length representation $h_k^s$. $h_k^g$ and $h_k^s$ are concatenated in order to obtain the recipe representation $h_k^r$. On the image side, the image encoder simply produces the fixed-length representation $h_k^v$. Then, the recipe and image representations are mapped into the joint embedding space as: $\phi^r = W^r h_k^r + b^r$ and $\phi^v = W^v h_k^v + b^v$, respectively. Note that $W^r$ and $W^v$ are embedding matrices which are also learned. Finally, the complete model is trained end-to-end with positive and negative recipe-image pairs $(\phi^r,\phi^v)$ using the cosine similarity loss with margin defined as follows:
\[
	L_{cos}(\phi^r,\phi^v,y) = 
\scriptsize\begin{cases}
1 - \cos(\phi^r,\phi^v), & \mbox{if } y = 1 \\
\max(0, \cos(\phi^r,\phi^v) - \alpha), & \mbox{if } y = -1
\end{cases}
\]
where $\cos(.)$ is the normalized cosine similarity and $\alpha$ is the margin.


\section{Semantic Regularization}
\label{sec:semantic}



We incorporate additional regularization on our embedding through solving the same high-level classification problem in multiple modalities with shared high-level weights. We refer to this method as semantic regularization. The key idea is that if high-level discriminative weights are shared, then both of the modalities (recipe and image embeddings) should utilize these weights in a similar way which brings another level of alignment based on discrimination. 
We optimize this objective together with our joint embedding loss. Essentially the model also learns to classify any image or recipe embedding into one of the food-related semantic categories. We limit the effect of semantic regularization as it is not the main problem that we aim to solve. 

\smallskip\noindent{\bf Semantic Categories.} We start by assigning Food-101 categories to those recipes that contain them in their title. However, after this procedure we are only able to annotate $13\%$ of our dataset, which we argue is not enough labeled data for a good regularization. Hence, we compose a larger set of semantic categories purely extracted from recipe titles. We first obtain the top 2,000 most frequent bigrams in recipe titles from our training set. We manually remove those that contain unwanted characters (e.g., \emph{n'}, \emph{!}, \emph{?} or \emph{\&}) and those that do not have discriminative food properties (e.g., \emph{best pizza}, \emph{super easy} or \emph{5 minutes}). We then assign each of the remaining bigrams as the semantic category to all recipes that include it in their title. By using bigrams and Food-101 categories together we obtain a total of 1,047 categories, which cover $50\%$ of the dataset. \emph{chicken salad}, \emph{grilled vegetable}, \emph{chocolate cake} and \emph{fried fish} are some examples among the categories we collect using this procedure. All those recipes without a semantic category are assigned to an additional \emph{background} class. Although there is some overlap in the generated categories, $73\%$ of the recipes in our dataset (excluding those in the $background$ class) belong to a single category (i.e., only one of the generated classes appears in their title). For recipes where two or more categories appear in the title, the category with highest frequency rate in the dataset is chosen. 

\smallskip\noindent{\bf Classification.} To incorporate semantic regularization to the joint embedding, we use a single fully connected layer. Given the embeddings $\phi^{v}$ and $\phi^{r}$, class probabilities are obtained with $p_{r} = W^{c}\phi^{r}$ and $p_{v} = W^{c}\phi^{v}$ followed by a softmax activation. $W^{c}$ is the matrix of learned weights, which are shared between image and recipe embeddings to promote semantic alignment between them. Formally, we express the semantic regularization loss as $L_{reg}(\phi^{r},\phi^{v},c_r,c_v)$ where $c_r$ and $c_v$ are the semantic category labels for recipe and image, respectively. Note that $c_r$ and $c_v$ are the same if $(\phi^{r},\phi^{v})$ is a positive pair. Then, we can write the final objective as:

\begin{equation*}
\begin{aligned}
L(\phi^r,\phi^v,c_{r},c_{v},y) &=& L_{cos}(\phi^r,\phi^v,y) + \\
 & & \lambda L_{reg}(\phi^r,\phi^v,c_{r},c_{v}) 
\end{aligned}
\label{eqn:obj}
\end{equation*}

\smallskip\noindent{\bf Optimization}. We follow a two-stage optimization procedure while learning the model.
If we update both the recipe encoding and image network at the same time, optimization becomes oscillatory and even divergent. Previous work on cross-modality training \cite{castrejon2016learning,aytar2016cross} suggests training models for different modalities separately and fine tuning them jointly afterwards to allow alignment. 
Following this insight, we adopt a similar procedure when training our model. 
We first fix the weights of the image network, which are found from pre-training on the ImageNet object classification task, and learn the recipe encodings. This way the recipe network learns to align itself to the image representations and also learns semantic regularization parameters ($W^c$). Then we freeze the recipe encoding and semantic regularization weights, and learn the image network. This two-stage process is crucial for successful optimization of the objective function. After this initial alignment stage, we release all the weights to be learned. However, the results do not change much in this final, joint optimization. 
We take a step further from \cite{ASalvador:CVPR17} in our extended study and change the validation procedure to use median rank (MedR) score as our performance measure, like in \cite{CarvalhoM:SIGIR18}, while reimplementing our source code in PyTorch.
This strategy appears to be more stable than using the validation loss.


\smallskip\noindent{\bf Implementation Details.} All the neural network models are implemented using Torch7\footnote{\url{http://torch.ch/}} and PyTorch\footnote{\url{https://pytorch.org/}} frameworks. The margin $\alpha$ is selected as $0.1$ in joint neural embedding models. The regularization hyper-parameter is set as $\lambda = 0.02$ in all our experiments. While optimizing the cosine loss, we pick a positive recipe-image pairs with $20\%$ probability and a random negative recipe-image pair with $80\%$ probability from the training set.

\begin{table*}[!t]
  \centering
    \caption{\textcolor{black}{\textbf{Im2recipe retrieval comparisons on \recipe{}}}. Median ranks and recall rate at top $K$ are reported for baselines and our method. Note that the joint neural embedding models consistently outperform all the baseline methods.}
    \begin{tabular}{lrrrrrrrr}
      \toprule
      & \multicolumn{4}{c}{im2recipe} & \multicolumn{4}{c}{recipe2im}\\
      \cmidrule(lr){2-5} \cmidrule(lr){6-9}
      \multicolumn{1}{c}{} & medR & R@1 & R@5 & R@10 &  medR & R@1 & R@5 & R@10 \\
      \midrule
      random ranking 		   	               					       & 500 & 0.001 & 0.005 & 0.01 & 500  & 0.001 & 0.005 & 0.01 \\
      CCA \scriptsize{w/ skip-thoughts + word2vec (GoogleNews) + image features}    & 25.2& 0.11 & 0.26 & 0.35 & 37.0 &  0.07 & 0.20 & 0.29 \\
      CCA  \scriptsize{w/ skip-instructions + ingredient word2vec + image features} & 15.7 & 0.14 & 0.32 & 0.43 & 24.8 &  0.09 & 0.24 & 0.35 \\
      \midrule
      joint emb. only 	  										   & 7.2 & 0.20 & 0.45 & 0.58 & 6.9 & 0.20 & 0.46 & 0.58 \\
      joint emb. + semantic 										   & 5.2 & 0.24 & 0.51 & 0.65 & 5.1 & 0.25 & 0.52 & 0.65 \\
      \midrule
      \textcolor{black}{attention + SR.~\cite{ChenJ:ACMMMM18}}
      & \textcolor{black}{4.6} & \textcolor{black}{0.26} & \textcolor{black}{0.54} & \textcolor{black}{0.67} & \textcolor{black}{4.6} & \textcolor{black}{0.26} & \textcolor{black}{0.54} & \textcolor{black}{0.67}
      \\
      \textcolor{black}{AdaMine~\cite{CarvalhoM:SIGIR18}}
      & \textcolor{black}{1.0} & \textcolor{black}{0.40} & \textcolor{black}{0.69} & \textcolor{black}{0.77} & \textcolor{black}{1.0} & \textcolor{black}{0.40} & \textcolor{black}{0.68} & \textcolor{black}{0.79}
      \\         
      \bottomrule
    \end{tabular}
  \label{tab:im2recipe_results}
\end{table*}


\begin{figure}[!t]
\centering
\includegraphics[width=\columnwidth]{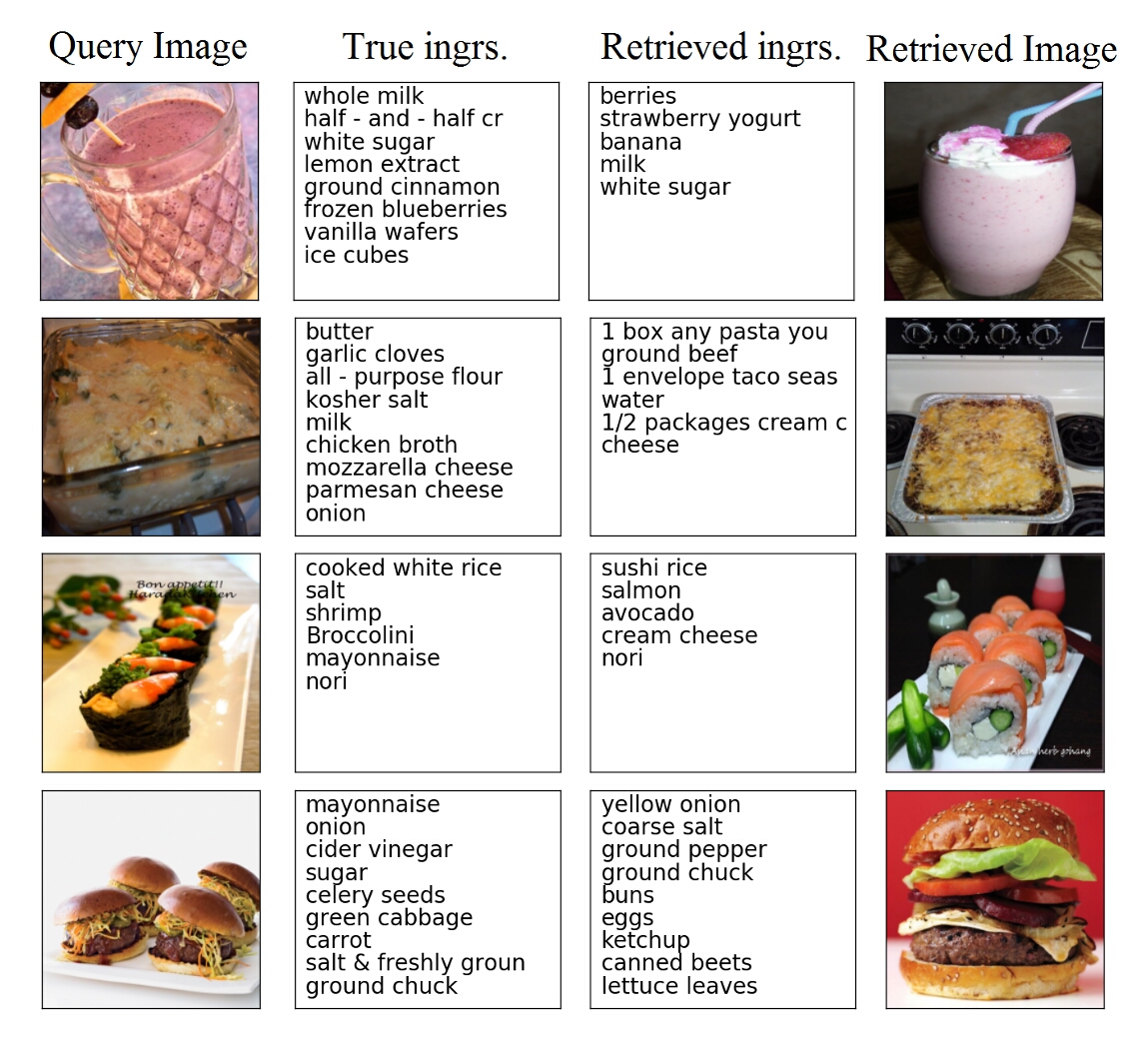}
\caption{\textbf{Im2recipe retrieval examples.} From left to right: (1) the query image, (2) its associated ingredient list, (3) the retrieved ingredients, and (4) the image associated to the retrieved recipe.}
\label{fig:rankings_i2r}
\end{figure}

The  models in Torch7 are trained on 4 NVIDIA Titan X with 12GB of memory for three days. The models in PyTorch are trained on 4 NVIDIA GTX 1080 with 8GB of memory for two and a half days (using a bigger batch size, i.e., 256 pairs instead of 150). When using \recipeplus, the training in PyTorch tends to take over a week, using a batch size of 256. For efficiency purposes, we store the recipe text part of the dataset in LMDB\footnote{\url{https://lmdb.readthedocs.io/en/release/}} format and load the images on the fly using \texttt{DataLoader} function of the PyTorch library. This way our PyTorch code does not require as much RAM as our Torch7 code does. As a side note, between the two reference libraries, we did experience that PyTorch in general uses less GPU memory.

\section{Experiments}
\label{sec:experiments}

We begin with the evaluation of our learned embeddings for the im2recipe retrieval task on the initial (i.e., recipe-website-only) version of our dataset (i.e., \recipe{}). Specifically, we study the effect of each component of our model and compare our final system against human performance for the im2recipe retrieval task. Then, using the best model architecture trained on the recipe-website-only version of the dataset, we compare its retrieval performance with the same one trained on the extended version of the dataset (i.e., \recipeplus{}) to evaluate the benefit of data extension through an image search engine. We further evaluate the two models on Food-101 dataset to assess their generalization ability. Finally, we analyze the properties of our learned embeddings through unit visualizations and explore different vector arithmetics in the embedding space on both the initial (\recipe{}) and the extended (\recipeplus{}) datasets.

\subsection{Im2recipe Retrieval}
\label{ssec:experiments_i2r}

We evaluate all the recipe representations for im2recipe retrieval. Given a food image, the task is to retrieve its recipe from a collection of test recipes. We also perform recipe2im retrieval using the same setting. All results are reported for the test set.

\smallskip\noindent{\bf Comparison with the Baselines.} Canonical Correlation Analysis (CCA) is one of the strongest statistical models for learning joint embeddings for different feature spaces when paired data are provided. We use CCA over many high-level recipe and image representations as our baseline. These CCA embeddings are learned using recipe-image pairs from the training data. In each recipe, the ingredients are represented with the mean word2vec across all its ingredients in the manner of \cite{le2014distributed}. The cooking instructions are represented with mean skip-thoughts vectors~\cite{skipthoughts} across the cooking instructions. A recipe is then represented as concatenation of these two features. We also evaluate CCA over mean ingredient word2vec and skip-instructions features as another baseline. The image features utilized in the CCA baselines are the ResNet-50 features before the softmax layer. Although they are learned for visual object categorization tasks on ImageNet dataset, these features are widely adopted by the computer vision community, and they have been shown to generalize well to different visual recognition tasks \cite{donahue2013decaf}.

For evaluation, given a test query image, we use cosine similarity in the common space for ranking the relevant recipes and perform im2recipe retrieval. The recipe2im retrieval setting is evaluated likewise. We adopt the test procedure from image2caption retrieval task \cite{karpathy2015deep,vinyals2015show}. We report results on a subset of randomly selected 1,000 recipe-image pairs from the test set. We repeat the experiments 10 times and report the mean results. We report median rank (MedR), and recall rate at top $K$ (R@K) for all the retrieval experiments. To clarify, R@5 in the im2recipe task represents the percentage of all the image queries where the corresponding recipe is retrieved in the top $5$, hence higher is better.
 The quantitative results for im2recipe retrieval are shown in Table~\ref{tab:im2recipe_results}. 

Our model outperforms the CCA baselines in all measures. As expected, CCA over ingredient word2vec and skip-instructions perform better than CCA over word2vec trained on GoogleNews~\cite{word2vec} and skip-thoughts vectors that are learned over a large-scale book corpus~\cite{skipthoughts}. In $65\%$ of all evaluated queries, our method can retrieve the correct recipe given a food image. The semantic regularization notably improves the quality of our embedding for im2recipe task which is quantified with the medR drop from $7.2$ to $5.2$ in Table~\ref{tab:im2recipe_results}. The results for recipe2im task are also similar to those in the im2recipe retrieval setting.

\textcolor{black}{Table~\ref{tab:im2recipe_results} also presents results originally reported in~\cite{ChenJ:ACMMMM18} and~\cite{CarvalhoM:SIGIR18} on \recipe{}. Attention-based modeling of \cite{ChenJ:ACMMMM18} achieves slight performance increases whereas double-triplet learning scheme of \cite{CarvalhoM:SIGIR18} leads to larger performance gains in both retrieval settings. 
}

Fig.~\ref{fig:rankings_i2r} compares the ingredients from the original recipes (true recipes) with the retrieved recipes (coupled with their corresponding image) for different image queries. 
As can be observed in Fig.~\ref{fig:rankings_i2r}, our embeddings generalize well and allow overall satisfactory recipe retrieval results. However, at the ingredient level, one can find that in some cases our model retrieves recipes with missing ingredients. This usually occurs due to the lack of fine-grained features (e.g., confusion between \emph{shrimps} and \emph{salmon}) or simply because the ingredients are not visible in the query image (e.g., \emph{blueberries} in a \emph{smoothie} or \emph{beef} in a \emph{lasagna}).

\begin{table*}[!t]
  \centering
    \caption{\textcolor{black}{\textbf{Ablation studies on \recipe{}.}} Effect of the different model components to the median rank, medR (the lower is better).}
  \begin{tabular}{llrrrrrr}
  \toprule
     & \multirow{2}{*}{Joint emb. methods} & \multicolumn{3}{c}{im2recipe} & \multicolumn{3}{c}{recipe2im}\\
     \cmidrule(lr){3-5} \cmidrule(lr){6-8}
     &  & medR-1K&medR-5K&medR-10K& medR-1K& medR-5K&medR-10K \\
    \midrule
	\multirow{3}{*}{{VGG-16}}     & fixed vision       & 15.3 & 71.8 & 143.6 & 16.4 & 76.8 &  152.8 \\
    									 & finetuning (ft)    & 12.1 & 56.1 & 111.4 & 10.5 & 51.0 &  101.4 \\
                                         & ft + semantic reg. & 8.2  & 36.4 &  72.4 & 7.3  & 33.4 &  64.9  \\
    \midrule
	\multirow{3}{*}{{ResNet-50}}  & fixed vision       & 7.9 & 35.7 & 71.2 & 9.3 & 41.9 &  83.1  \\
    									 & finetuning (ft)    & 7.2 & 31.5 & 62.8 & 6.9 & 29.8 &  58.8  \\
                                         & ft + semantic reg. & 5.2 & 21.2 & 41.9 & 5.1 & 20.2 &  39.2  \\
    \bottomrule
  \end{tabular}
  \label{tab:im2recipe_ablation}
\end{table*}

\begin{table*}[!t]
 \centering
  \caption{\textcolor{black}{\textbf{Comparison with human performance on im2recipe task on \recipe{}.}} The mean results are highlighted as bold for better visualization. Note that on average our method with semantic regularization performs better than average AMT worker.}
   \resizebox{\textwidth}{!}{%
   \begin{tabular}{lrrrrrrrrrrrrrrrr}
     \toprule
 \multicolumn{1}{c}{} & all recipes & \multicolumn{6}{c}{course-specific recipes}  & \multicolumn{9}{c}{dish-specific recipes} \\ 
     \cmidrule(lr){3-8}
     \cmidrule(lr){9-17}
 \multicolumn{1}{c}{} &  & { dessert}  & { salad}  & { bread}  & { beverage}  & { soup-stew} & { course-mean} & { pasta}  & { pizza}  & { steak}  & { salmon}  & { smoothie}  & { hamburger}  & { ravioli}  & { sushi}  & { dish-mean} \\ 
 \midrule
     { human} & {\bf  81.6 $\pm$ 8.9} & 52.0 & 70.0 & 34.0 & 58.0 & 56.0 &  {\bf 54.0  $\pm$  13.0} & 54.0 & 48.0 & 58.0 & 52.0 & 48.0 & 46.0 & 54.0 & 58.0 & {\bf 52.2  $\pm$  04.6}   \\ 
 { joint-emb. only} & {\bf 83.6  $\pm$  3.0} & 76.0 & 68.0 & 38.0 & 24.0 & 62.0 & {\bf 53.6  $\pm$  21.8} & 58.0 & 58.0 & 58.0 & 64.0 & 38.0 & 58.0 & 62.0 & 42.0 & {\bf 54.8 $\pm$  09.4}  \\ 
 { joint-emb.+semantic} & {\bf 84.8  $\pm$  2.7} & 74.0 & 82.0 & 56.0 & 30.0 & 62.0 & {\bf 60.8  $\pm$  20.0} & 52.0 & 60.0 & 62.0 & 68.0 & 42.0 & 68.0 & 62.0 &  44.0 & {\bf 57.2 $\pm$  10.1}  \\ 
 \bottomrule
 \end{tabular}
 }
 \label{tab:human}
\end{table*}

\begin{table*}[!t]
  \centering
    \caption{\textcolor{black}{\textbf{Comparison between models trained on \recipe{} vs.\ \recipeplus{}.}} Median ranks and recall rate at top $K$ are reported for both models. They have similar performance on the \recipe{} test set in terms of medR and R@K. However, when testing on the \recipeplus{} test set, the model trained on \recipeplus{} yields significantly better medR and better R@5 and R@10 scores. \textcolor{black}{In this table, \recipe{} refers to the \textit{intersection} dataset.}}
    \begin{tabular}{lrrrrrrrr}
      \toprule
      & \multicolumn{4}{c}{\recipe{} test set} & \multicolumn{4}{c}{\recipeplus{} test set}\\
      \midrule
      & \multicolumn{8}{c}{im2recipe}\\
      \cmidrule(lr){2-5} \cmidrule(lr){6-9}
      \multicolumn{1}{c}{} & medR & R@1 & R@5 & R@10 &  medR & R@1 & R@5 & R@10 \\
      \midrule
      \recipe{} training set & 5.1 & 0.24 & 0.52 & 0.64 & 13.6 & 0.15 & 0.35 & 0.46 \\
      \recipeplus{} training set & 5.7 & 0.21 & 0.49 & 0.62 & 8.6  & 0.17 & 0.42 & 0.54 \\
      \midrule
      & \multicolumn{8}{c}{recipe2im}\\
      \cmidrule(lr){2-5} \cmidrule(lr){6-9}
      \multicolumn{1}{c}{} & medR & R@1 & R@5 & R@10 &  medR & R@1 & R@5 & R@10 \\      
      \midrule
      \recipe{} training set & 4.8 & 0.27 & 0.54 & 0.65 & 11.9 & 0.17 & 0.38 & 0.48 \\
      \recipeplus{} training set & 4.6  & 0.26 & 0.54 & 0.66 & 6.8  & 0.21 & 0.46 & 0.58 \\
      \bottomrule
    \end{tabular}
  \label{tab:extended_vs_initial}
\end{table*}

\smallskip\noindent{\bf Ablation Studies.} We also analyze the effect of each component in our our model in several optimization stages. The results are reported in Table~\ref{tab:im2recipe_ablation}. Note that here we also report medR with $1K$, $5K$ and $10K$ random selections to show how the results scale in larger retrieval problems. As expected, visual features from the ResNet-50 model show a substantial improvement in retrieval performance when compared to VGG-16 features. Even with ``fixed vision" networks the joint embedding achieved $7.9$ medR using ResNet-50 architecture. Further ``fine-tuning'' of vision networks slightly improves the results. Although it becomes a lot harder to decrease the medR in small numbers, additional ``semantic regularization" improves the medR in both cases.

\smallskip\noindent{\bf Comparison with Human Performance.}
In order to better assess the quality of our embeddings we also evaluate the performance of humans on the im2recipe task. The experiments are performed through AMT. For quality purposes, we require each AMT worker to have at least $97\%$ approval rate and have performed at least 500 tasks before our experiment. In a single evaluation batch, we first randomly choose $10$ recipes and their corresponding images. We then ask an AMT worker to choose the correct recipe, out of the $10$ provided recipes, for the given food image. This multiple choice selection task is performed $10$ times for each food image in the batch. The accuracy of an evaluation batch is defined as the percentage of image queries correctly assigned to their corresponding recipe.     

 The evaluations are performed for three levels of difficulty. The batches (of $10$ recipes) are randomly chosen from either all the test recipes (easy), recipes sharing the same course (e.g.,\ soup, salad, or beverage; medium), or recipes sharing the name of the dish (e.g.,\ salmon, pizza, or ravioli; hard). As expected--for our model as well as the AMT workers--the accuracies decrease as tasks become more specific. In both coarse and fine-grained tests, our method performs comparably to or better than the AMT workers. As hypothesized, semantic regularization further improves the results (see Table~\ref{tab:human}).

In the ``all recipes'' condition, $25$ random evaluation batches ($25 \times 10$ individual tasks in total) are selected from the entire test set. Joint embedding with semantic regularization performs the best with $3.2$ percentage points improvement over average human accuracy. For the course-specific tests, $5$ batches are randomly selected within each given meal course. Although, on average, our joint embedding's performance is slightly lower than the humans', with semantic regularization our joint embedding surpasses humans' performance by $6.8$ percentage points. In dish-specific tests, five random batches are selected if they have the dish name (e.g.,\ pizza) in their title. With slightly lower accuracies in general, dish-specific results also show similar behavior. Particularly for the ``beverage" and ``smoothie" results, human performance is better than our method, possibly because detailed analysis is needed to elicit the homogenized ingredients in drinks. Similar behavior is also observed for the ``sushi" results where fine-grained features of the sushi roll's center are crucial to identify the correct sushi recipe.

\begin{figure*}[!t]
 \centering
 \includegraphics[width=\textwidth]{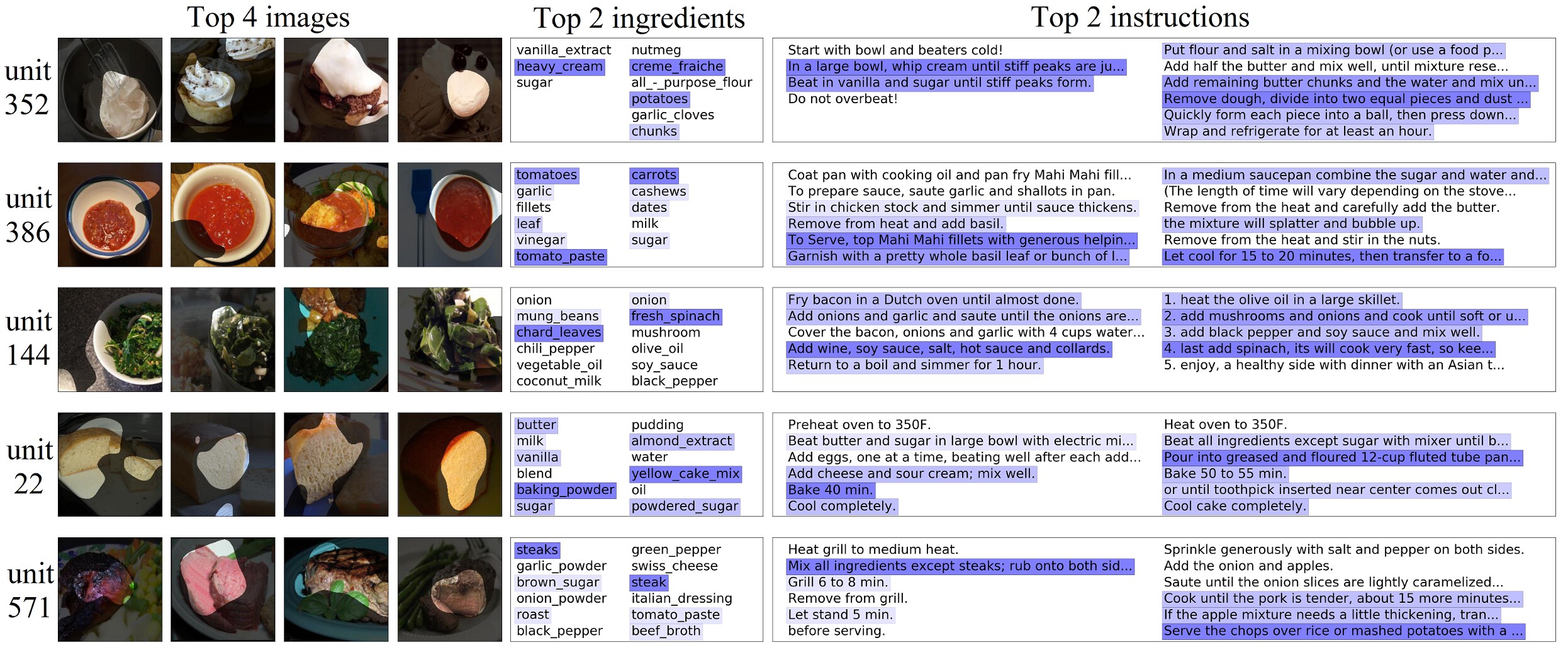}
 \caption{\textbf{Localized unit activations.} We find that ingredient detectors emerge in different units in our embeddings, which are aligned across modalities (e.g., unit 352: ``cream'', unit 22: ``sponge cake'' or unit 571: ``steak'').}
 \label{fig:occlusion}
\end{figure*}

\smallskip\noindent{\bf \recipe{} vs.\ \recipeplus{} Comparison}. One of the main questions of the current study is how beneficial it is to incorporate images coming from a Web search engine into the initial collection of images obtained from recipe websites. One way to assess this is to compare im2recipe retrieval performance of a network architecture trained on \recipe{} with im2recipe retrieval performance of the same network architecture trained on \recipeplus{}. In Table~\ref{tab:extended_vs_initial}, we present im2recipe retrieval results achieved on both test sets. As can be seen, there is a clear benefit when we evaluate both models on the \recipeplus{} test set. The model trained on \recipeplus{} obtains a significantly better medR, 5 points lower in both retrieval tasks, and higher R@5 and R@10, in some cases up to a 10 percentage point increase. When looking into the \recipe{} test set, both models perform similarly. These results clearly demonstrate the benefit of using external search engines to extend the imagery content of \recipe{}. Note that the retrieval results on Tables \ref{tab:im2recipe_results} and \ref{tab:extended_vs_initial} slightly differ due to the fact that we use a modified version of the dataset (see \textit{intersection} dataset in Table \ref{tvtdatastats}) in the latter experiment. As we explained earlier in Section~\ref{sec:dataset}, this is done mainly to have a fair comparison of im2recipe retrieval results on both versions of the dataset.

\begin{table}[!t]
  \centering
    \caption{\textbf{Im2recipe retrieval comparisons on Food-101 dataset.} Median ranks and recall rate at top $K$ are reported for both models. Note that the model trained on \recipeplus{} performs better than the model trained on \recipe{}. \textcolor{black}{In this table, \recipe{} refers to the \textit{intersection} dataset.}}
    \begin{tabular}{lrrrr}
      \toprule
      & \multicolumn{4}{c}{im2recipe} \\
      \cmidrule(lr){2-5}
      \multicolumn{1}{c}{} & medR & R@1 & R@5 & R@10 \\
      \midrule
      \recipe{} training set 	& 17.35 & 16.13 & 33.68 & 42.53 \\
      \midrule
      \recipeplus{} training set & 10.15 & 21.89 & 42.31 & 51.14 \\
      \midrule
      & \multicolumn{4}{c}{recipe2im} \\
      \cmidrule(lr){2-5}
      \recipe{} training set 	& 4.75 & 26.19 & 54.52 & 67.50 \\
      \midrule
      \recipeplus{} training set & 2.60 & 37.38 & 65.00 & 76.31 \\
      \bottomrule
    \end{tabular}

  \label{tab:food101_results}
\end{table}

\smallskip\noindent{\bf Model Generalization Ability Comparison}. We experiment further to evaluate whether \recipeplus{} dataset improves the performance of our model on other food image datasets. For this purpose, we evaluate both our trained models on the popular Food-101 dataset~\cite{bossard2014food}. The Food-101 dataset is a classification dataset containing 101 food categories and 1,000 images for each one of these 101 food categories, totaling up to 101,000 images.

Our method of evaluation involves randomly sampling an image and a recipe corresponding to each of the Food-101 categories. The images are taken from the Food-101 dataset, while the recipes are taken from the test partition of the \textit{intersection} dataset. Here, a recipe is considered to belong to a category only if the recipe title string matches with the Food-101 category. Here, we only sample images and recipes from those categories that correspond to at least N recipes among the test recipes that we sample from.

After sampling an image and a corresponding recipe for each category that is common enough, we evaluate our models on the retrieval task. In the im2recipe direction, we provide our model with the image and expect it to retrieve the corresponding recipe. In the recipe2im direction, we provide our model with the recipe and expect it to retrieve the corresponding image. We show the retrieval results of both models in Table~\ref{tab:food101_results}. Note that the model trained on \recipeplus{} consistently outperforms the model trained on \recipe{}.

One possible explanation for \recipeplus{} dataset giving an advantage on the Food-101 task is that there might be an overlap between the images used to train the model on the \recipeplus{} and the Food-101 images. Further, it is possible that there might be images in \recipeplus{} training set that overlap with the Food-101 dataset that are not in the initial training set. This would give the model trained on \recipeplus{} an unfair advantage. We perform the following procedure to test whether this is true. First, we feed in all of the images in the \recipeplus{} training set and the Food-101 images into an 18 layer residual network that was pre-trained on ImageNet. The network outputs a prediction vector for each of these images. We next note that if an image in the extended training set has an exact copy in the Food-101 dataset, then both images must have the same prediction vector. When checking the prediction vectors of the images in Food-101 and the \recipeplus{} training set, we did not find any overlapping prediction vectors, meaning that the images between Food-101 and \recipeplus{} training set do not overlap.

\subsection{Analysis of the Learned Embedding}
\label{ssec:experiments_analysis}

To gain further insight into our neural embedding, we perform a series of qualitative analysis experiments. We explore whether any semantic concepts emerge in the neuron activations and whether the embedding space has certain arithmetic properties.

\begin{figure*}[ht!]
\vspace{2em}

\subfloat{
\includegraphics[width=0.48\textwidth]{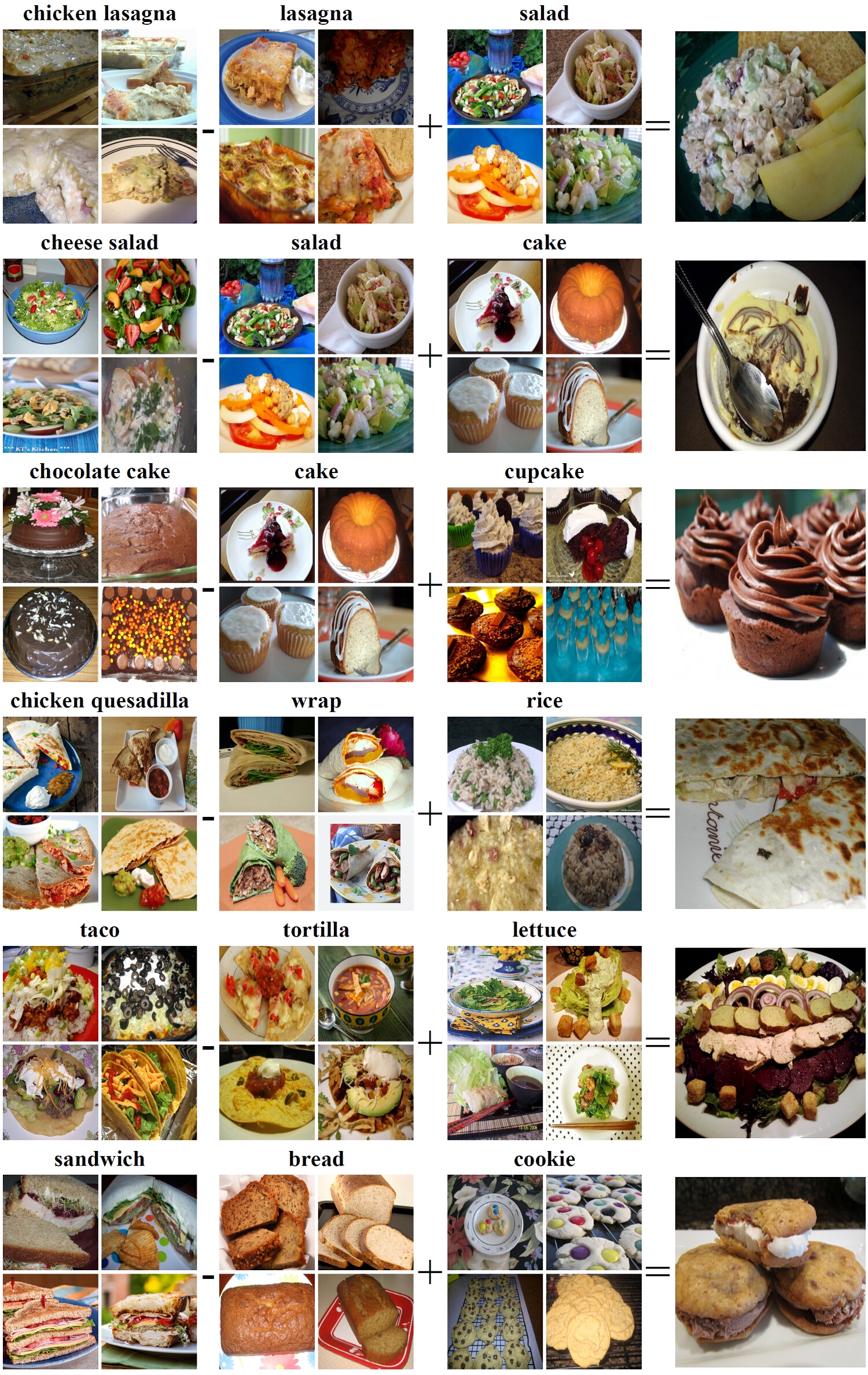}
}
\rulesep
\rulesep
\subfloat{
\includegraphics[width=0.48\textwidth]{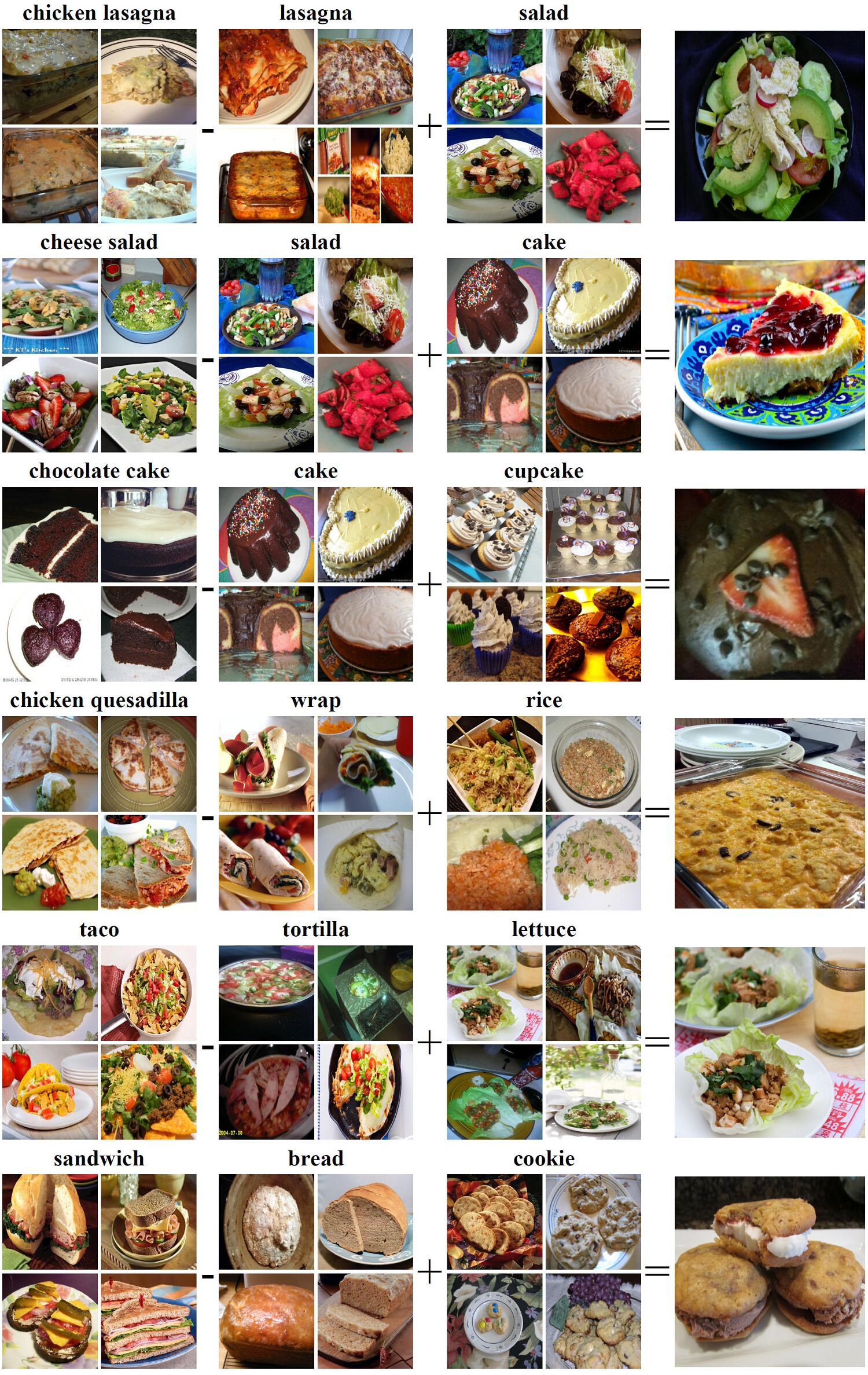}
}
\caption{\textbf{Analogy arithmetic results using recipe embeddings} on the \recipe{} test set. On the left hand side are arithmetic results using the model trained on \recipe{}. On the right hand side are the arithmetic results for the model trained on \recipeplus{}. We represent the average vector of a query with the images from its 4 nearest neighbors. In the case of the arithmetic result, we show the nearest neighbor only.}
\label{fig:analogy_rec}
\end{figure*}

\smallskip\noindent{\bf Neuron Visualizations.} Through neural activation visualization, we investigate if any semantic concepts emerge in the neurons in our embedding vector despite not being explicitly trained for that purpose. We pick the top activating images, ingredient lists, and cooking instructions for a given neuron. Then we use the methodology introduced by Zhou et al.~\cite{unitvis} to visualize image regions that contribute the most to the activation of specific units in our learned visual embeddings. We apply the same procedure on the recipe side to also obtain those ingredients and recipe instructions to which certain units react the most. Fig.~\ref{fig:occlusion} shows the results for the same unit in both the image and recipe embedding. We find that certain units display localized semantic alignment between the embeddings of the two modalities.

\begin{figure*}[ht!]
\vspace{2em}
\label{fig:analogy_image}

\subfloat{
\includegraphics[width=0.48\textwidth]{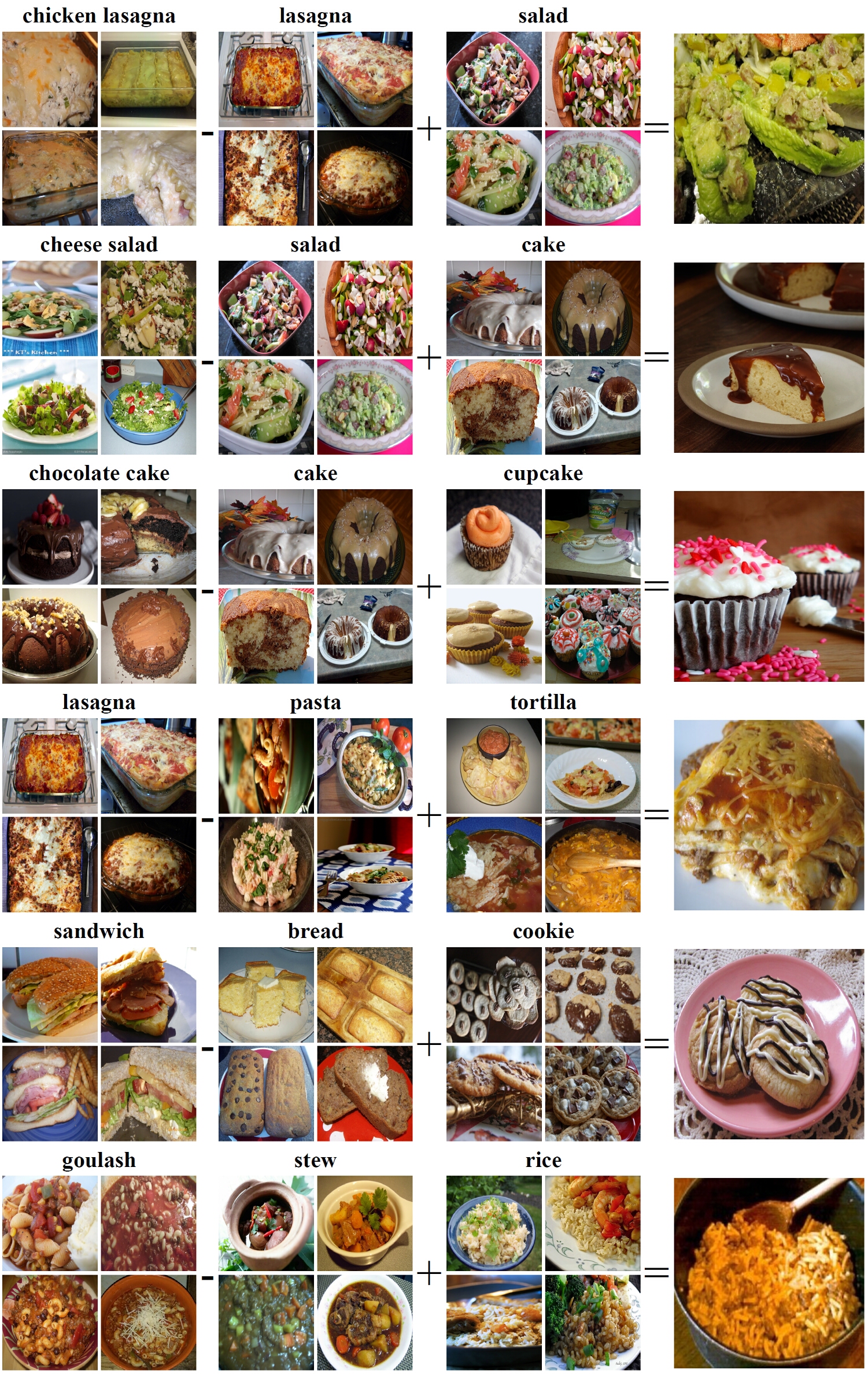}
}
\rulesep
\rulesep
\subfloat{
\includegraphics[width=0.48\textwidth]{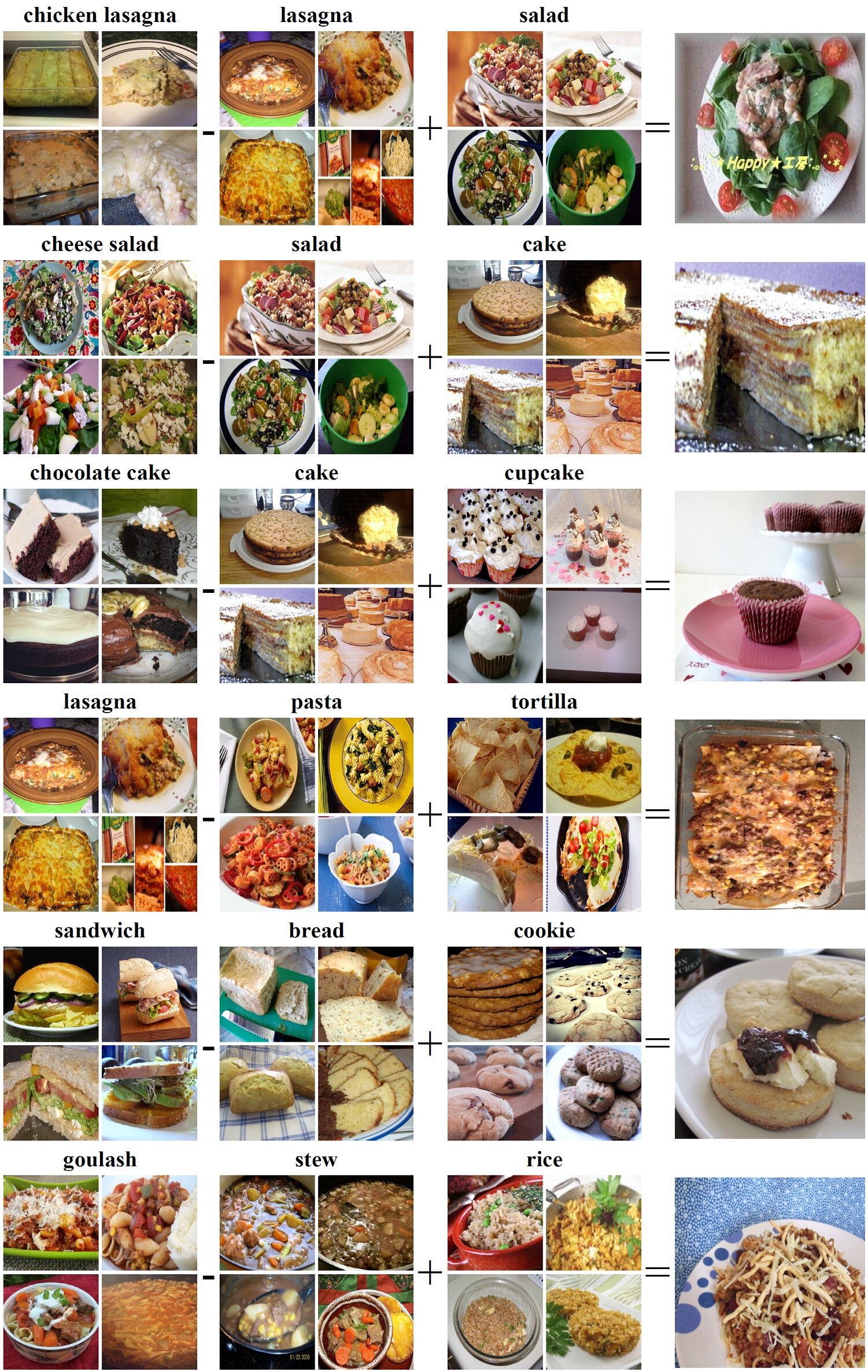}
}
\caption{\textbf{Analogy arithmetic results using image embeddings} on the \recipe{} test set. On the left hand side are arithmetic results using the model trained on \recipe{}. On the right hand side are the arithmetic results for the model trained on \recipeplus{}. We represent the average vector of a query with the images from its four nearest neighbors. In the case of the arithmetic result, we show the nearest neighbor only.}
\label{fig:analogy_im}
\end{figure*}

\begin{figure*}[ht!]
\vspace{2em}
\includegraphics[width=\textwidth]{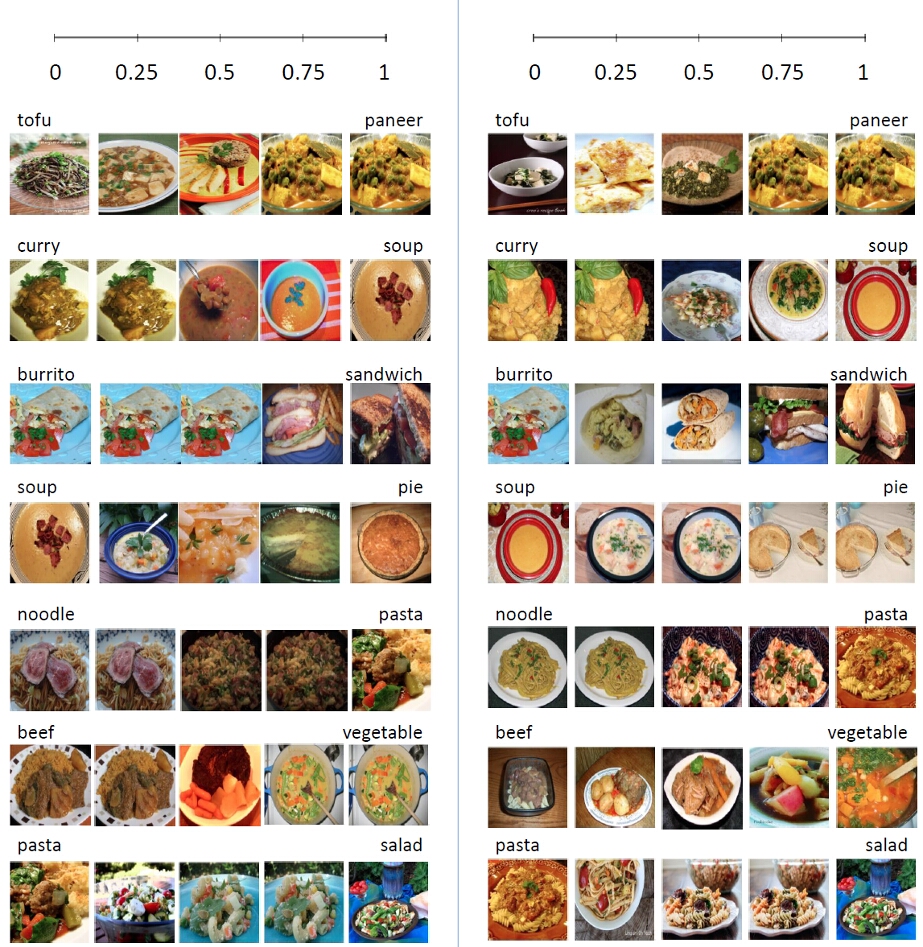}
\caption{\textbf{Fractional arithmetic results using recipe embeddings} on the \recipe{} test set. On the left hand side are arithmetic results using the model trained on \recipe{}. On the right hand side are the arithmetic results for the model trained on \recipeplus{}. For each model, we fractionally interpolate across two example concepts (for instance, ``salad" and ``pasta"). We find the retrieved results for $x \times v($``concept 1"$) + (1-x) \times v($``concept 2"$)$, where $x$ varies from $0$ to $1$.}
\label{fig:fractional_rec}
\end{figure*}

\begin{figure*}[ht!]
\vspace{2em}
\includegraphics[width=\textwidth]{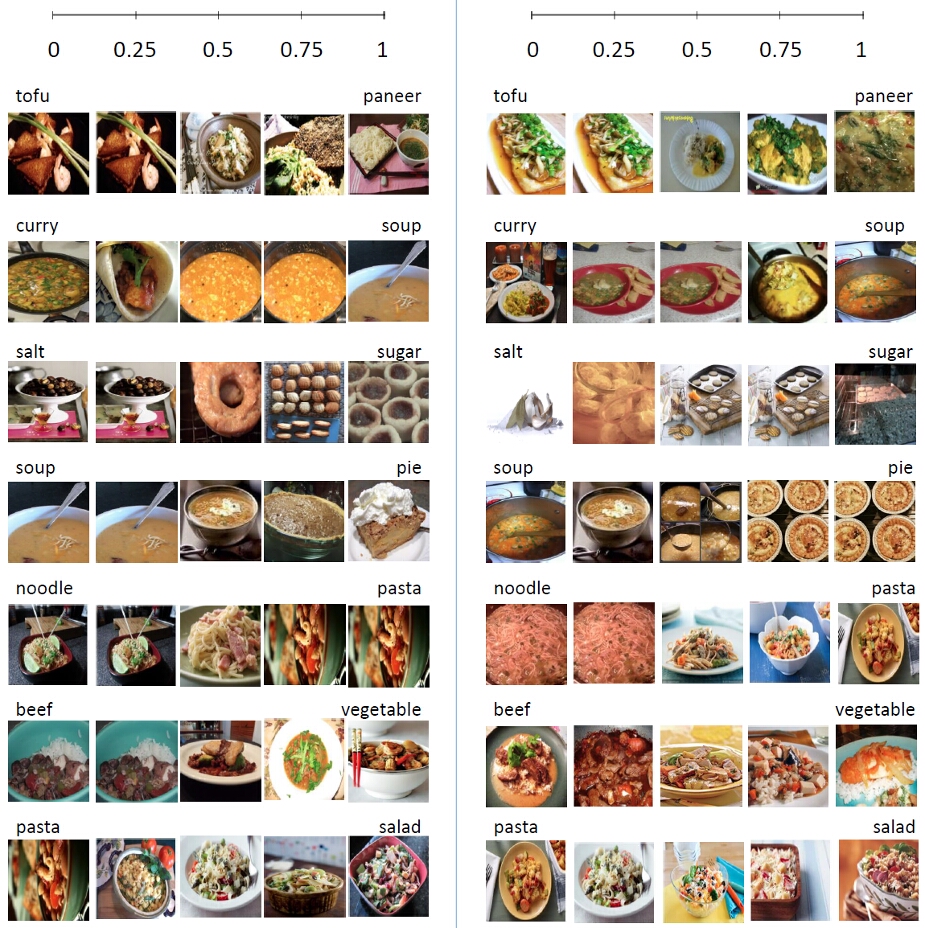}
\caption{\textbf{Fractional arithmetic results using image embeddings} on the \recipe{} test set. On the left hand side are arithmetic results using the model trained on \recipe{}. On the right hand side are the arithmetic results for the model trained on \recipeplus{}. For each model, we fractionally interpolate across two example concepts (for instance, ``salad" and ``pasta"). We find the retrieved results for $x \times v($``concept 1"$) + (1-x) \times v($``concept 2"$)$, where $x$ varies from $0$ to $1$.}
\label{fig:fractional_im}
\end{figure*}

\smallskip\noindent{\bf Semantic Vector Arithmetic.} Different works in the literature \cite{word2vec,arithm2} have used simple arithmetic operations to demonstrate the capabilities of their learned representations. In the context of food recipes, one would expect that $v($``chicken pizza''$) - v($``pizza''$) + v($``salad''$) = v($``chicken salad''$)$, where $v$ represents the map into the embedding space. We demonstrate that our learned embeddings have such properties by applying the previous equation template to the averaged vectors of recipes that contain the queried words in their title. We apply this procedure in the recipe and image embedding spaces and show results in Fig.~\ref{fig:analogy_rec} and Fig.~\ref{fig:analogy_im}, respectively. Our findings suggest that the learned embeddings have semantic properties that translate to simple geometric transformations in the learned space. Furthermore, the model trained on \recipeplus{} is better able to capture these semantic properties in the embedding space. The improvement is most seriously observable on the recipe arithmetic. Among the recipe analogy examples, notice that the result for the \recipeplus{} dataset for ``chicken quesadilla" - ``wrap" + ``rice" returns a casserole dish, while for the \recipe{} dataset we have a quesadilla dish. The casserole dish is much closer to matching the ``chicken rice" result that we expect in this instance. Additionally, note how ``taco" - ``tortilla" + ``lettuce" returns a salad for the \recipe{} model and a lettuce wrap for the \recipeplus{} model. Here, the former model is likely doing arithmetic over the ingredients in the dish - a taco without tortilla likely comprises of a salad, into which lettuce is added to give a salad-like dish. On the other hand, the \recipeplus{} model does arithmetic over higher level semantic concepts - it returns a lettuce wrap, which is the closest analogue to a taco which has the tortilla substituted out with lettuce. We can thus see how the \recipeplus{} model has a greater ability to capture semantic concepts in the recipe embedding space, and also performs somewhat better in general. If we examine the results of both models for the analogy task with image embeddings, then the \recipeplus{} model shows less of an improvement in general. However, we can still see differences between the two models. For instance, if we examine the ``taco" - ``tortilla" + ``lettuce" analogy, then the \recipe{} model returns a result in which the lettuce is mixed in with other ingredients to form a salad. However, the \recipeplus{} model returns a result in which a salad is placed on top of a large piece of lettuce. This result is similar in a way to the lettuce wrap result, as the piece of lettuce is not just mixed in with the other ingredients, but acts as more of an object into which other ingredients are placed. All in all, the \recipeplus{} training set allows our model to better capture high level semantic concepts.

\smallskip\noindent\textbf{Fractional Arithmetic.} Another type of arithmetic we examine is fractional arithmetic, in which our model interpolates across the vector representations of two concepts in the embedding space. Specifically, we examine the results for $x \times v($``concept 1"$) + (1-x) \times v($``concept 2"$)$, where $x$ varies from $0$ to $1$. We expect this to have interesting applications in spanning the space across two food concepts, such as pasta and salad, by adjusting the value of $x$ to make the dish more ``pasta-like" or ``salad-like" for example. We apply this procedure in the recipe and image embedding spaces and show results in Fig.~\ref{fig:fractional_rec} and Fig.~\ref{fig:fractional_im}, respectively. With both fractional image arithmetic and fractional recipe arithmetic, we hope that adjusting the fractional coefficient will allow us to explore more fine-grained combinations of two concepts. However, the results are often not so fine-grained. For instance, in the ``burrito" and ``sandwich" example for the model trained on the \recipe{} dataset for recipe fractional arithmetic, choosing a burrito coefficient of $0$ does not yield different results from choosing the coefficient to be $0.5$. Note that on the other hand, the model trained on the \recipeplus{} dataset is able to provide distinct results for each fractional coefficient value for this example. In general though, both models are able to effectively explore the gradient of recipes or food images between two different food concepts. For instance, note the models' results for the ``curry" and ``soup" examples, in both the image and recipe modalities. The most ``curry-like" image tends to have some broth, but is much chunkier than the images. As we increase the coefficient of ``soup", we see the food becoming less chunky and more broth-like. Such examples reflect the ability of our model to explore the space between food concepts in general.

The results of our fractional arithmetic experiments suggest that the recipe and image embeddings learned in our model are semantically aligned, which broaches the possibility of applications in recipe modification (e.g., ingredient replacement, calorie adjustment) or even cross-modal generation.

\section{Conclusion}
\label{sec:conclusion}

In this paper, we present \recipeplus{}, the largest structured recipe dataset to date, the im2recipe problem, and neural embedding models with semantic regularization which achieve impressive results for the im2recipe task. \textcolor{black}{The experiments conducted using AMT, together with the fact that on the \recipe{} test set we obtain the same test performance using \recipeplus{}, show that the extended dataset is not much noisier. Moreover, the fact that this expansion strategy greatly helps on the Food 101 dataset demonstrates the value for generalizability.} Additionally, we explored the properties of the resulting recipe and food representations by evaluating different vector arithmetics on the learned embeddings, which hinted at the possibility of applications such as recipe modification or even cross-modal recipe generation. 

More generally, the methods presented here could be gainfully applied to other ``recipes'' like assembly instructions, tutorials, and industrial processes. Further, we hope that our contributions will support the creation of automated tools for food and recipe understanding and open doors for many less explored aspects of learning such as compositional creativity and predicting visual outcomes of action sequences. 

\ifCLASSOPTIONcompsoc
  \section*{Acknowledgments}
\else
  \section*{Acknowledgment}
\fi

This work was supported by CSAIL-QCRI collaboration project. 

\ifCLASSOPTIONcaptionsoff
  \newpage
\fi



\bibliographystyle{IEEEtran}
\bibliography{egbib,IEEEabrv}

\begin{IEEEbiography}[{\includegraphics[width=1in,height=1.25in,clip,keepaspectratio]{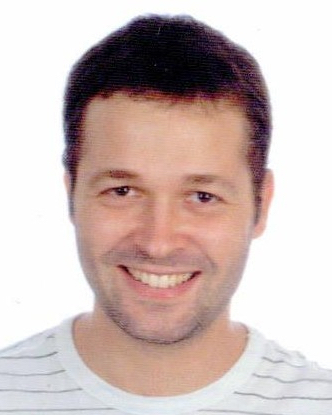}}]{Javier Mar\'in} received the B.Sc. degree in Mathematics at the Universitat de les Illes
Balears in 2007. In June 2013 he received his Ph.D. in computer vision at the Universitat Aut\'onoma de Barcelona. In 2017 he was a postdoctoral research associate at the Massachusetts Institute of Technology (MIT). Before that, he worked as an algorithm development engineer in the automotive sector, and as a researcher and project manager in both neuroscience and space fields. He currently combines working in the private sector as a senior data scientist at Satellogic Solutions with being a research affiliate at MIT. His research interests lie mainly in the area of computer vision and machine learning, focusing recently in cross-modal learning, object recognition and semantic segmentation.
\end{IEEEbiography}
\vspace{-5mm}
\begin{IEEEbiography}[{\includegraphics[width=1in,height=1.25in,clip,keepaspectratio]{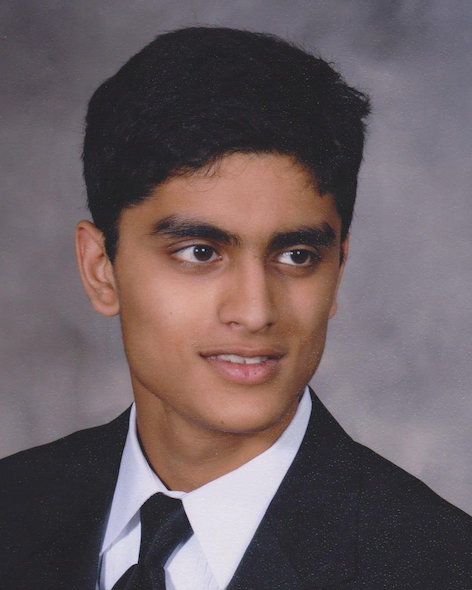}}]{Aritro Biswas} received a Bachelor's degree in Computer Science at the Massachusetts Institute of Technology (MIT). He received his Master's degree in Computer Science at MIT. Recently, his research has focused on using computer vision for two applications: (i) understanding the content of food images and (ii) disaster recognition for images of humanitarian disasters.
\end{IEEEbiography}
\vspace{-5mm}
\begin{IEEEbiography}[{\includegraphics[width=1in,height=1.25in,clip,keepaspectratio]{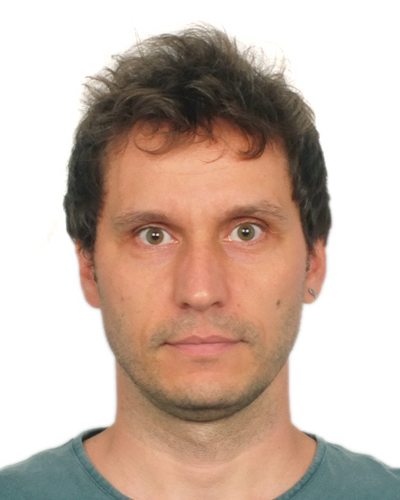}}]{Ferda Ofli} (S'07--M'11--SM'18) received the B.Sc. degrees both in electrical and electronics engineering and computer engineering, and the Ph.D. degree in electrical engineering from Koc University, Istanbul, Turkey, in 2005 and 2010, respectively. From 2010 to 2014, he was a Postdoctoral Researcher at the University of California, Berkeley, CA, USA. He is currently a Scientist at the Qatar Computing Research Institute (QCRI), part of Hamad Bin Khalifa University (HBKU). His research interests cover computer vision, machine learning, and multimedia signal processing.
He is an IEEE and ACM senior member with over 45 publications in refereed conferences and journals including CVPR, WACV, TMM, JBHI, and JVCI. He won the Elsevier JVCI best paper award in 2015, and IEEE SIU best student paper award in 2011. He also received the Graduate Studies Excellence Award in 2010 for outstanding academic achievement at Koc University.
\end{IEEEbiography}
\vspace{-5mm}
\begin{IEEEbiography}[{\includegraphics[width=1in,height=1.25in,clip,keepaspectratio]{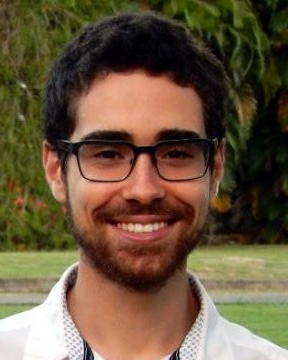}}]{Nicholas Hynes} is a graduate student at the University of California Berkeley and research scientist at Oasis Labs. His research interests are generally in the domain of efficient machine learning on shared private data.\end{IEEEbiography}

\begin{IEEEbiography}[{\includegraphics[width=1in,height=1.25in,clip,keepaspectratio]{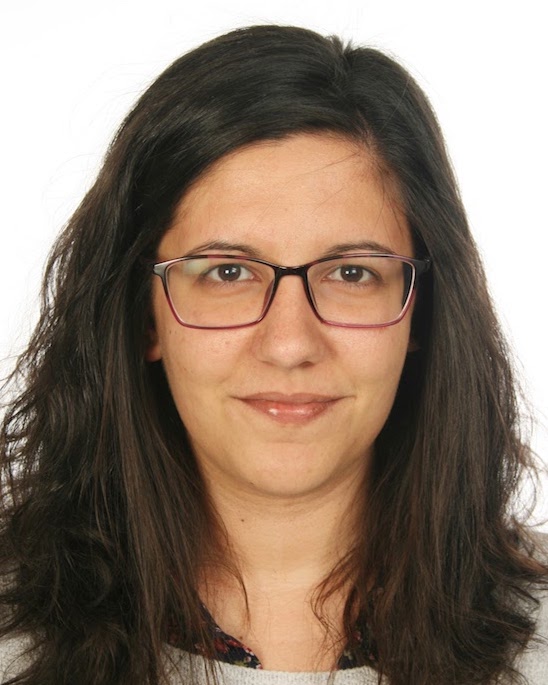}}]{Amaia Salvador}  is a PhD candidate at Universitat Politècnica de Catalunya under the advisement of Professor Xavier Giró and Professor Ferran Marqués. She obtained her B.S. in Audiovisual Systems Engineering from Universitat Politècnica de Catalunya in 2013, after completing her thesis on interactive object segmentation at the ENSEEIHT Engineering School in Toulouse. She holds a M.S. in Computer Vision from Universitat Autònoma de Barcelona. She spent the summer of 2014 at the Insight Centre for Data Analytics in the Dublin City University, where she worked on her master thesis on visual instance retrieval. In 2015 and 2016 she was a visiting student at the National Institute of Informatics and the Massachusetts Institute of Technology, respectively. During the summer of 2018, she interned at Facebook AI Research in Montreal.\end{IEEEbiography}
\vspace{-5mm}
\begin{IEEEbiography}[{\includegraphics[width=1in,height=1.25in,clip,keepaspectratio]{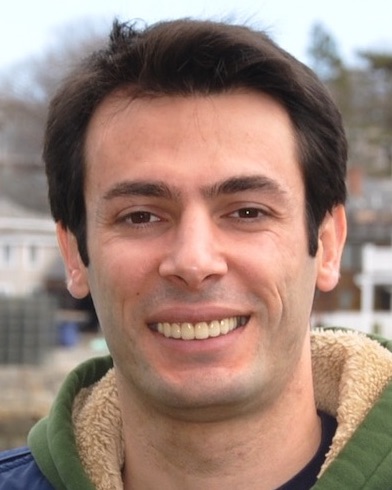}}]{Yusuf Aytar} is a Research Scientist at DeepMind since July 2017. He was a post-doctoral research associate at Massachusetts Institute of Technology (MIT) between 2014-2017. He received his D.Phil. degree from University of Oxford. As a Fulbright scholar, he obtained his M.Sc. degree from University of Central Florida (UCF), and his B.E. degree in Computer Engineering in Ege University. His research is mainly concentrated on computer vision, machine learning, and transfer learning.\end{IEEEbiography}
\vspace{-5mm}
\begin{IEEEbiography}[{\includegraphics[width=1in,height=1.25in,clip,keepaspectratio]{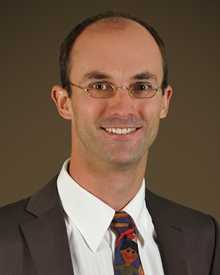}}]{Ingmar Weber} is the Research Director for Social Computing at the Qatar Computing Research Institute (QCRI). As an undergraduate Ingmar studied mathematics at Cambridge University, before pursuing a PhD at the Max-Planck Institute for Computer Science. He subsequently held positions at the Ecole Polytechnique Federale de Lausanne (EPFL) and Yahoo Research Barcelona, as well as a visiting researcher position at Microsoft Research Cambridge. He is an ACM, IEEE and AAAI senior member.
\end{IEEEbiography}
\vspace{-5mm}
\begin{IEEEbiography}[{\includegraphics[width=1in,height=1.25in,clip,keepaspectratio]{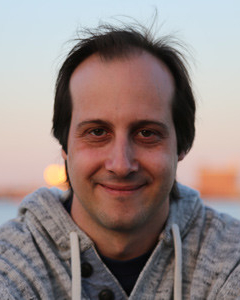}}]{Antonio Torralba} received the degree in telecommunications engineering from Telecom BCN, Barcelona, Spain, in 1994 and the Ph.D. degree in signal, image, and speech processing from the Institut National Polytechnique de Grenoble, France, in 2000. From 2000 to 2005, he spent postdoctoral training at the Brain and Cognitive  Science  Department  and  the  Computer Science and Artificial Intelligence Laboratory, MIT. He is now a Professor of Electrical Engineering and Computer Science at the Massachusetts Institute of Technology (MIT). Prof. Torralba is an Associate Editor of the International Journal in Computer Vision, and has served as program chair for the Computer Vision and Pattern Recognition conference in 2015. He received the 2008 National Science Foundation (NSF) Career award, the best student paper award at the IEEE Conference on Computer Vision and Pattern Recognition (CVPR) in 2009, and the 2010 J. K. Aggarwal Prize from the International Association for Pattern Recognition (IAPR).
\end{IEEEbiography}







\end{document}